\pdfoutput=1

\documentclass[11pt]{article}

\usepackage[final]{acl}

\usepackage{times}
\usepackage{latexsym}
\usepackage{booktabs}
\usepackage{array}
\usepackage{makecell}
\usepackage{pifont}
\usepackage{multirow}
\usepackage{bbding}
\usepackage{xcolor}
\usepackage{enumitem}
\usepackage[T1]{fontenc}

\usepackage[utf8]{inputenc}

\usepackage{microtype}

\usepackage{inconsolata}

\usepackage{graphicx}

\title{IntentionESC: An Intention-Centered Framework for Enhancing Emotional Support in Dialogue Systems}

\author{\textbf{Xinjie Zhang$^1$, Wenxuan Wang$^1$, Qin Jin${^{1,\dagger}}$} \\
  \small $^1$School of Information, Renmin University of China \\
  {\small \tt \{zhangxinjie827, wangwenxuan, qjin\}@ruc.edu.cn}\\
}

\begin{document}
\maketitle
\def\thefootnote{$\dagger$}\footnotetext{Corresponding Author.}

\begin{abstract}

In emotional support conversations, unclear intentions can lead supporters to employ inappropriate strategies, inadvertently imposing their expectations or solutions on the seeker. Clearly defined intentions are essential for guiding both the supporter’s motivations and the overall emotional support process.
In this paper, we propose the \textbf{Intention}-centered \textbf{E}motional \textbf{S}upport \textbf{C}onversation (\textbf{IntentionESC}) framework, which defines the possible intentions of supporters in emotional support conversations, identifies key emotional state aspects for inferring these intentions, and maps them to appropriate support strategies. 
While Large Language Models (LLMs) excel in text generating, they fundamentally operate as probabilistic models trained on extensive datasets, lacking a true understanding of human thought processes and intentions. To address this limitation, we introduce the \textbf{I}ntention \textbf{CE}ntric \textbf{C}hain-\textbf{o}f-\textbf{T}hought (\textbf{ICECoT}) mechanism. ICECoT enables LLMs to mimic human reasoning by analyzing emotional states, inferring intentions, and selecting suitable support strategies, thereby generating more effective emotional support responses. To train the model with ICECoT and integrate expert knowledge, we design an automated annotation pipeline that produces high-quality training data. Furthermore, we develop a comprehensive evaluation scheme to assess emotional support efficacy and conduct extensive experiments to validate our framework. Our data and code are available at \href{https://github.com/43zxj/IntentionESC_ICECoT}{https://github.com/43zxj/IntentionESC\_ICECoT}.

\end{abstract}
\section{Introduction}

\begin{figure}[ht]
\centering
\includegraphics[scale=0.5]{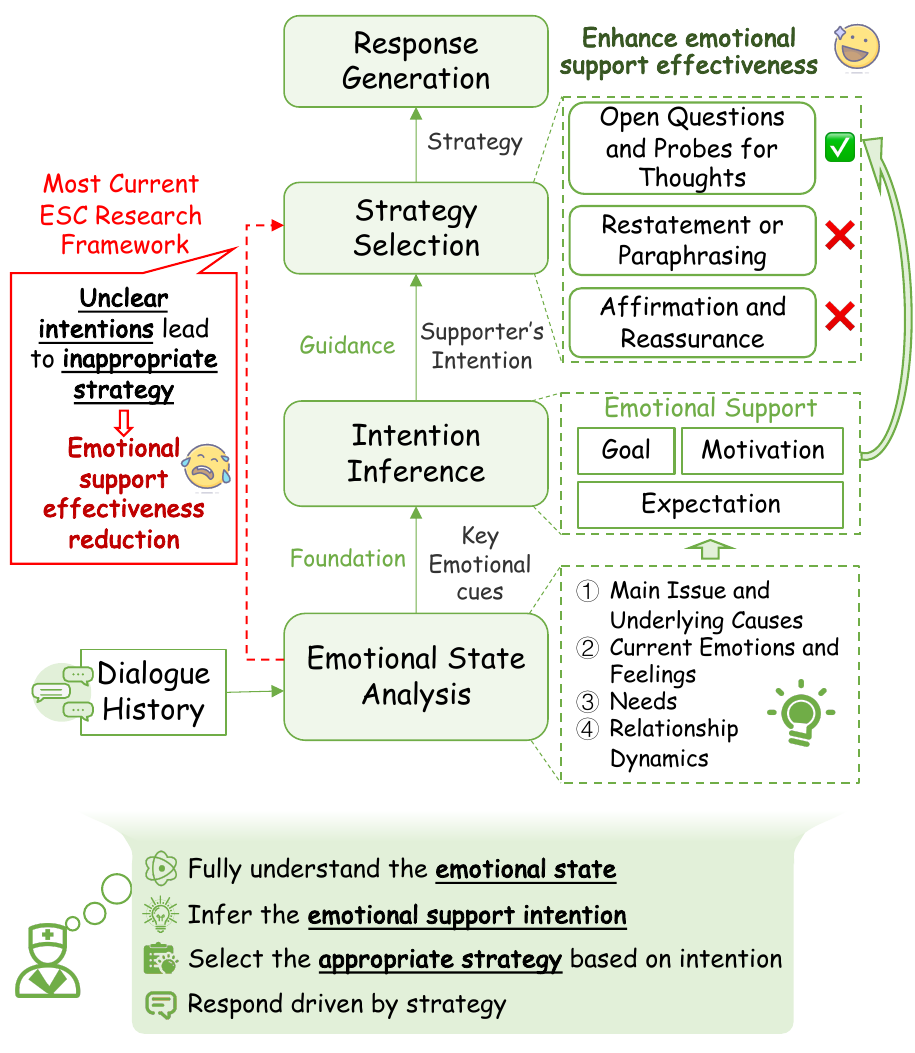}
\vspace{-8pt}
\caption{Illustration of our IntentionESC framework and ICECoT. The ICECoT mechanism involves four key reasoning steps. Guided by the IntentionESC framework, the system first analyzes the four key aspects of the emotional state, then infers the supporter’s emotional support intention—including goals, motivations, and expectations—which informs strategy selection and enables the generation of effective emotional support responses.}
\label{fig:Framework}
\end{figure}
Emotional support plays a crucial role in social interactions by helping individuals navigate emotional challenges through understanding and comfort~\cite {burleson2003emotional, cdata1987communicating, cutrona1987provisions}. Integrating this capability into dialogue systems enhances their ability to engage in more human-like and empathetic interactions, aligning better with users’ expectations and preferences~\cite{rains2020support}. However, delivering effective emotional support is a complex process requiring systematic strategies and procedures. Drawing on \textit{Helping Skills} Theory~\cite{hill2009helping}, we observe that a
 therapist's intention---the rationale behind selecting specific helping skills and the goal of counseling---plays a pivotal role in effective support. Likewise, in emotional support conversations, only when supporters act with a clear intention can they select suitable strategies and provide meaningful emotional support.

\textbf{Intention}, in this context, refers to a supporter’s internal motivations, goals, and expectations when providing support. Imagine a scenario where a seeker, after experiencing a major setback such as job loss, expresses emotional distress. 
If the supporter’s intention is to “help the seeker feel understood and comforted,” they might adopt strategies like \textit{Affirmation and Reassurance}, responding with, “I understand how hard this must be for you; it’s a huge blow for anyone.” Such responses effectively address the seeker's emotional needs. Conversely, if the supporter’s intention is to “solve the problem quickly”, they might provide direct advice like, “It’s okay, you can start looking for another job right away.” While practical, this approach may overlook the seeker's emotional state, potentially leading to ineffective or even counterproductive interactions. As \textit{Helping Skills} theory emphasizes, clarifying intention is essential before selecting a strategy ~\cite{hill2009helping}.

While existing research has explored the seekers' intentions in emotional support conversations (ESC)~\cite{GLHGijcai2022p600}, there has been limited focus on the supporters' intentions. \citet{ESConvliu2021towards} proposes the ESC framework for emotional support conversations, which involves three stages and various support strategies, addressing three sub-problems related to emotion states, strategies, and evaluation. 
However, most studies inspired by this framework have primarily focused on the “\textbf{what to do}” aspect---emotional state modeling and strategy selection---while neglecting the “\textbf{why}” aspect, specifically the underlying intentions driving emotional support. 

To bridge this gap, we propose the  \textbf{Intention-centered Emotional Support Conversation} (\textbf{IntentionESC}) framework. 
This framework defines supporters' intentions, identifies the emotional state preconditions necessary for inferring these intentions, and maps these intentions to appropriate strategies. Drawing from counseling practices~\cite{hill2001list}, we adapt and refine a list of 12 emotional support intentions suited for emotional support scenarios. 
By analyzing the situational information embedded in these intentions, we identify four key aspects of emotional state analysis crucial for intention inference.
Furthermore, by establishing explicit links between intentions and strategies, we outline specific strategies supporters can employ to acomplish each intention. 
As illustrated in Figure ~\ref{fig:Framework}, our {IntentionESC} framework provides better structured guidance for effective emotional support.

Large language models (LLMs) exhibit advanced conversational capabilities due to extensive pretraining on diverse text corpora, offering new opportunities for ESC~\cite{cheng2024cooper, zhang-etal-2024-escot}. However, LLMs, being fundamentally probabilistic, generate language based on learned patterns rather than genuine comprehension of human thought and intention. To align LLMs with the IntentionESC framework and enhance their ability to generate emotional support responses, we propose the \textbf{I}ntention \textbf{CE}ntric \textbf{C}hain-\textbf{o}f-\textbf{T}hought (\textbf{ICECoT}) mechanism. ICECoT operates in three stages: \textit{emotional state analysis} → \textit{intention inference} → \textit{strategy selection and response generation}. 
The \textit{emotional state analysis} stage forms the foundation for all subsequent stages by extracting crucial contextual information from the conversation.
Building on this, the \textit{intention inference} stage determines the supporters' intentions required to guide the selection of appropriate support \textit{strategies} and generation of \textit{responses}.
Unlike ESCoT~\cite{zhang-etal-2024-escot}, ICECoT links emotional state analysis and strategy selection through intention, creating a more coherent and interpretable response generation process. 
By making intention inference explicit, ICECoT enhances the transparency and trustworthiness of emotional support models. Additionally, we establish an automated annotation pipeline to infuse expert knowledge into data, effectively transferring human insights to LLMs. 

The main contributions of our work include: (1) We are the first to focus on supporters' intentions in emotional support conversations and propose the intention-centered emotional support conversation (IntentionESC) framework; (2) We introduce ICECoT, a novel reasoning chain mechanism that integrates emotional state analysis, intention inference, and strategy selection for emotional support response generation, alongside an automated annotation process for generating reasoning chain data; (3) We design a new evaluation framework to assess emotional support dialogue models and conduct extensive experiments demonstrating the effectiveness of intention-driven emotional support. 

\section{Related Work}

\begin{figure*}[ht]
\centering
\includegraphics[scale=0.4]{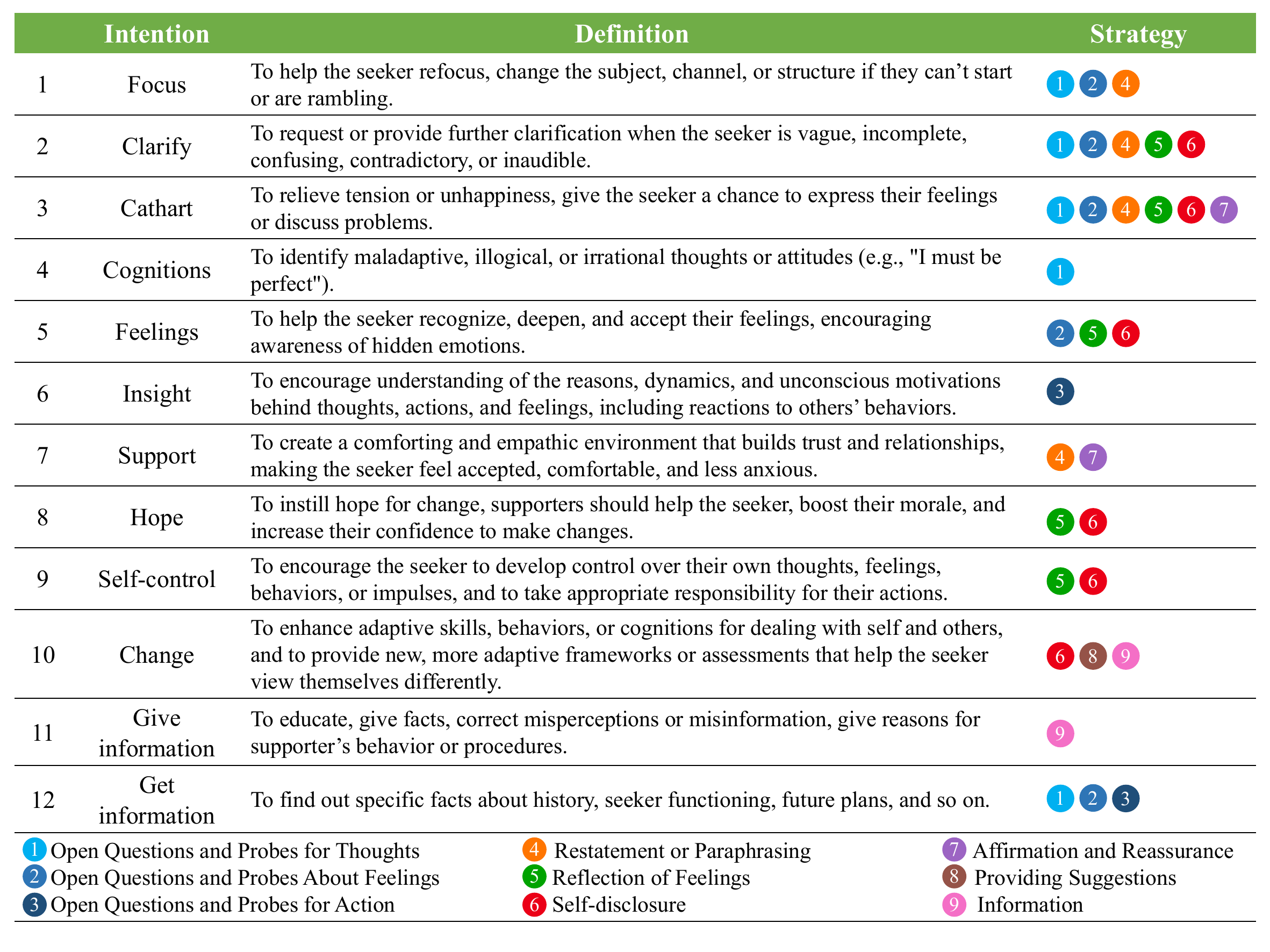}
\vspace{-15pt}
\caption{The list of emotional support intentions, definitions, and strategies.}
\label{fig:intentionFramework}
\end{figure*}

\paragraph{Emotional Support Conversation}

\citet{ESConvliu2021towards} proposes the ESConv dataset and several sub-tasks for emotional support conversations, including Support Strategy Selection and Strategy-Constrained Response Generation, Emotion State Modeling, and Evaluation of Support Effectiveness. Many studies are based on this framework to explore ESC tasks. 
Some work focuses on emotional state modeling. GLHG~\cite{GLHGijcai2022p600} proposes a Global-to-Local Hierarchical Graph Network to capture multi-source information, including the global cause that describes the emotional problem and the local intention that implies the current speaker’s state.
CauESC~\cite{chen2024cauesc} investigates emotion causes and the intra/internal effects of emotions, devising an attention mechanism to reason about the effects triggered by emotion causes.
ESCoT~\cite{zhang-etal-2024-escot} proposes an emotional support response inference method that focuses on the emotional process based on how emotions are generated.
Some research also focuses on support strategy selection and strategy-constrained response generation. \citet{tu2022misc} proposes the MIxed Strategy-Aware Model (MISC), a method for responding by skillfully blending strategies. MultiESC~\cite{MultiESCcheng2022improving} conducts support strategy planning by capturing subtle emotional expressions and causes, anticipating user feedback, and dynamically tracking the user’s emotional state.
Additionally, some studies focus on incorporating commonsense knowledge~\cite{Commonsensewang2023enhancing} and personality~\cite{cheng2023pal}.
A key link between emotional modeling and strategy selection lies in the clear understanding of the seeker’s contextual state, which guides both the choice of strategy and the generation of responses—the \textbf{intention}. However, existing studies have overlooked the specific intentions behind providing emotional support.


\section{IntentionESC Framework}

In emotional support conversations, intention serves as the guiding force behind effective strategies and responses. This section introduces our IntentionESC framework, which outlines the possible intentions of supporters, identifies key emotional state aspects necessary for inferring the intentions, and maps them to appropriate support strategies to enhance the quality of emotional support.

\subsection{Intentions in ESC}
\citet{hill2001list} compiled a comprehensive set of intentions from psychological counseling practices, refining and validating them through interactive analysis. 
These intentions help therapists and counselors facilitate emotional healing and personal growth for their clients.  Beyond professional therapy, these insights can also guide emotional support conversations in everyday contexts.

However, not all intentions from counseling directly apply to emotional support conversations. For example, the intention to “Set limits” is relevant in therapy, where structured boundaries (e.g., time or fees) exist, but less applicable in casual emotional support. Similarly, intentions such as “Reinforce change” or “Challenge” 
may be too complex or risky for supporters without professional training~\cite{ESConvliu2021towards}. To address these discrepancies, we refine and adapt the original list of intentions~\cite{hill2001list} to better suit ESC needs, resulting in  12 distinct emotional support intentions. These, summarized in Figure~\ref{fig:intentionFramework}, are tailored to reflect the motivations and expectations of supporters in ESC scenarios while incorporating expert insights. This structured framework provides clear guidance for selecting appropriate strategies, empowering supporters to deliver more effective emotional support.

\subsection{Aspects for Inferring Intentions}
A deep understanding of the seeker's emotional state is fundamental for forming appropriate intentions and providing meaningful support. While previous works have explored emotional states modeling~\cite{GLHGijcai2022p600, chen2024cauesc, zhang-etal-2024-escot}, intention inference requires a more nuanced and comprehensive approach. 
Helping intentions arise from situational understanding, where key triggers influence how supporters respond. 
For example, inferring the \textit{Focus} intention---defined as “To help a seeker refocus, change the subject, channel, or structure if they can’t start or are rambling”---requires the supporter to first understand the seeker's main issue. 
Based on professional insights and contextual analysis, we identify four key aspects of emotional state modeling for intention inference:
1) the seeker's main issue and its underlying causes, 2) the seeker's current emotions and feelings, 3) explicit or implicit needs expressed by the seeker, and 4) the relationship between the seeker and the supporter. 
Incorporating these key aspects enables supporters to infer intentions more accurately, considering the interaction context and situational nuances. 

\subsection{Intention-driven Strategy Selection}
In therapeutic settings, strategies (or helping skills) are selected based on the therapist's helping intention~\cite{hill2009helping}. Similarly, in ESC, supporters should first establish appropriate intentions before choosing specific strategies to ensure effective support. 
Drawing from Helping Skills~\cite{hill2009helping}, where the correspondence between intentions and strategies are outlined, we set up a mapping of each support intention to potential strategies, as shown in Figure~\ref{fig:intentionFramework}.
By clarifying the relationships between strategies and intentions, the IntentionESC framework enables supporters to align their intentions with appropriate strategies, enhancing the overall effectiveness of delivering emotional support.

\section{Response Generation Mechanism}
The IntentionESC framework provides structured guidance for emotional support, ensuring that responses align with the seeker's needs. Building on this foundation, we introduce the \textbf{Intention-Centric Chain-of-Thought ({ICECoT})} mechanism, which enables LLMs to follow a structured reasoning process akin to professional emotional support practices. 
ICECoT follows a four-step process: first understanding seeker's emotional state, then identifying the supporter's underlying intention, followed by choosing an appropriate support strategy, and finally producing a well-grounded emotional support response. 



\subsection{ICECoT Mechanism}



Professional emotional support follows a structured process: assessing the seeker's emotional state, determining an appropriate intention, selecting a support strategy, and then formulating a response. Our ICECoT mirrors this workflow to ensure contextually appropriate and intention-driven responses. Under ICECoT, the model first analyzes the seeker's emotional state, identifying their issues, emotions, and needs. Based on this understanding, it then infers supporter's intended goal, ensuring responses align with the seeker's actual needs rather than generic or misaligned strategies. The inferred intention guides strategy selection. Finally, the model generates a response that is both supportive and contextually appropriate.
This structured reasoning process not only improves the professionalism of generated responses but also enhances model interpretability by explicitly linking emotional understanding, intentions, and response strategies.

\begin{figure}[t]
\centering
\includegraphics[scale=0.3]{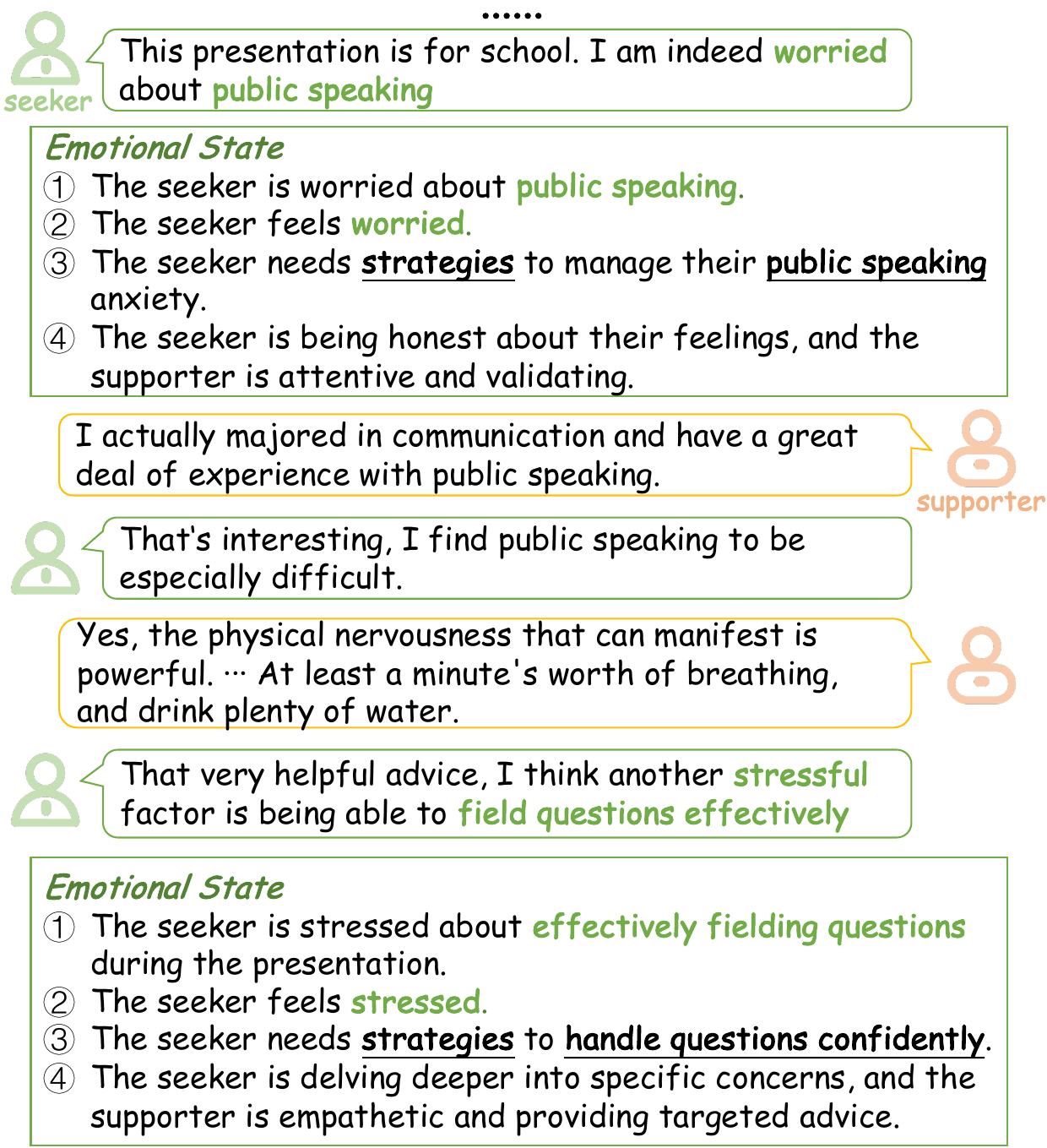}
\vspace{-6pt}
\caption{The case of emotional state annotation, which shows the content and changes of emotion state annotations throughout the dialogue process. The \ding{172}-\ding{175} correspond to the four key aspects of the emotional state.}
\label{fig:State}
\end{figure}

\subsection{Emotional State Analysis}

ICECoT employs four key aspects of emotional state analysis based on IntentionESC framework: (1) \textbf{Seeker’s Main Issue and Underlying Causes}: Identify the core issue and its root causes; (2) \textbf{Seeker’s Current Emotions and Feelings}: Capture explicit and implicit emotions; (3) \textbf{Seeker’s Needs}: Recognize desired support. including solutions  or assistance; (4) \textbf{Conversation Relationship Dynamics}: Monitor how the seeker-supporter relationship evolves.

Manual emotional state annotation is resource-intensive, so we leverage GPT-4 to automate this process efficiently on the ESConv dataset~\cite{ESConvliu2021towards}.
To capture the evolving emotional context, each seeker utterance is annotated with their cumulative emotional state, with all four aspects continuously updated as the conversation unfolds. 
This approach prevents redundant discussion of resolved issues, allows for detailed refinement of problem and cause descriptions based on seeker input, and facilitates the monitoring of dynamic changes in emotions, feelings, and relationship dynamics.
These guidelines are incorporated into the prompt, as detailed in Appendix~\ref{appd:Emotional_prompt}, enabling LLMs trained on this data to effectively master emotional state analysis.

\begin{figure}[t]
\centering
\includegraphics[scale=0.3]{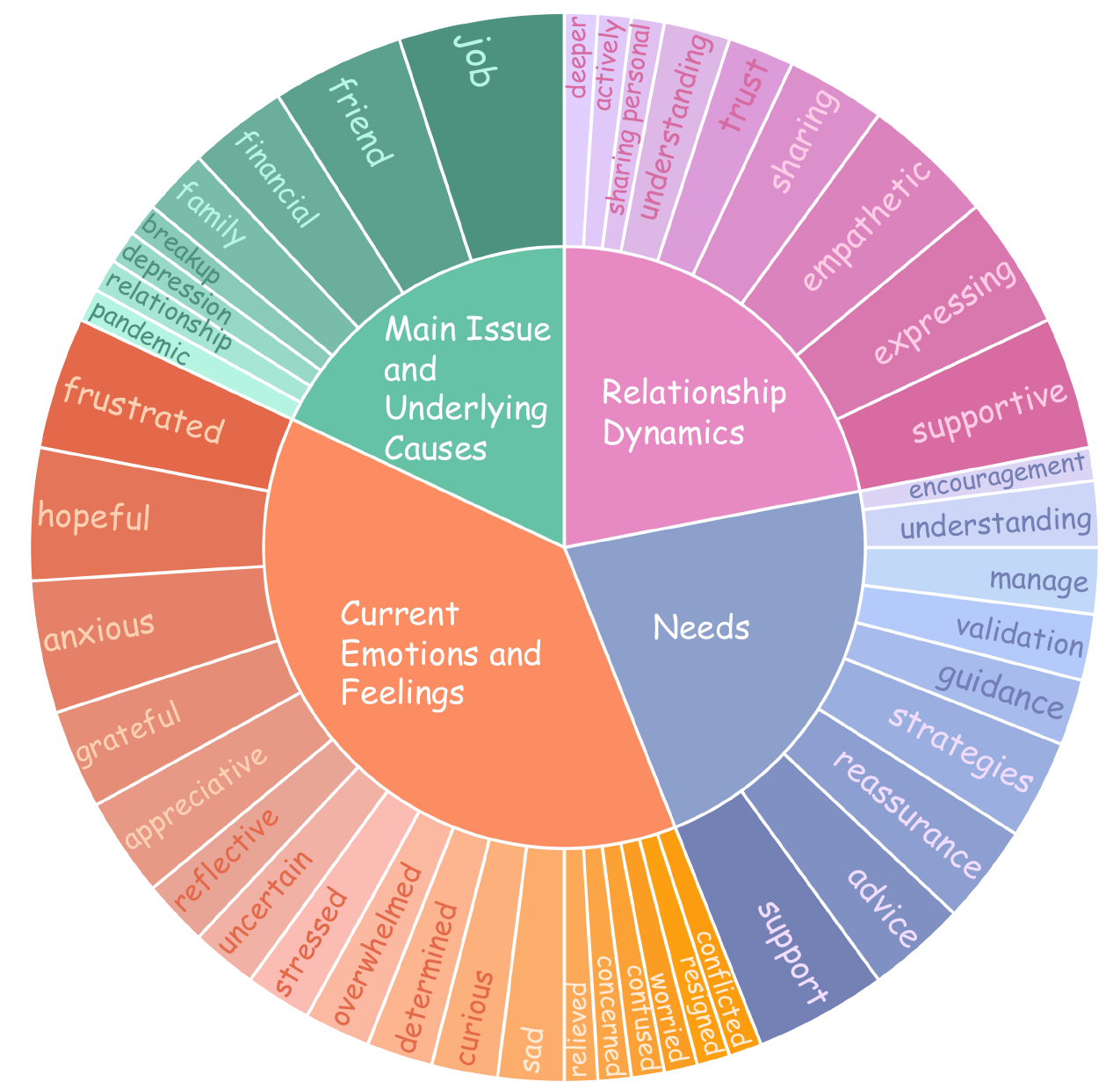}
\vspace{-6pt}
\caption{The keywords in emotional state. It shows the frequently occurring words in the annotations of four aspects, where larger and darker areas indicate higher frequencies.}
\label{fig:Kewords}
\end{figure}

Figure~\ref{fig:State} illustrates the generated annotations. 
The seeker's initial main issue is anxiety about giving a speech. As the conversation progresses, the supporter offers guidance, and new challenges surface, updating the seeker's issues and needs. The evolving relationship dynamics between the seeker and supporter reflect a deepening bond of trust. By analyzing the word frequency of the annotated content, we identify the most frequently appearing words in each aspect, as shown in Figure~\ref{fig:Kewords}. The keywords in the \textit{Main Issue and Underlying Causes} are almost identical to those identified in the original ESConv~\cite{ESConvliu2021towards} concerning the seeker’s problems. In addition to negative emotions, the seeker’s \textit{Current Emotions and Feelings} include positive words like `hopeful' and `grateful', reflecting the impact of emotional support. 

\begin{figure*}[ht]
\centering
\includegraphics[scale=0.45]{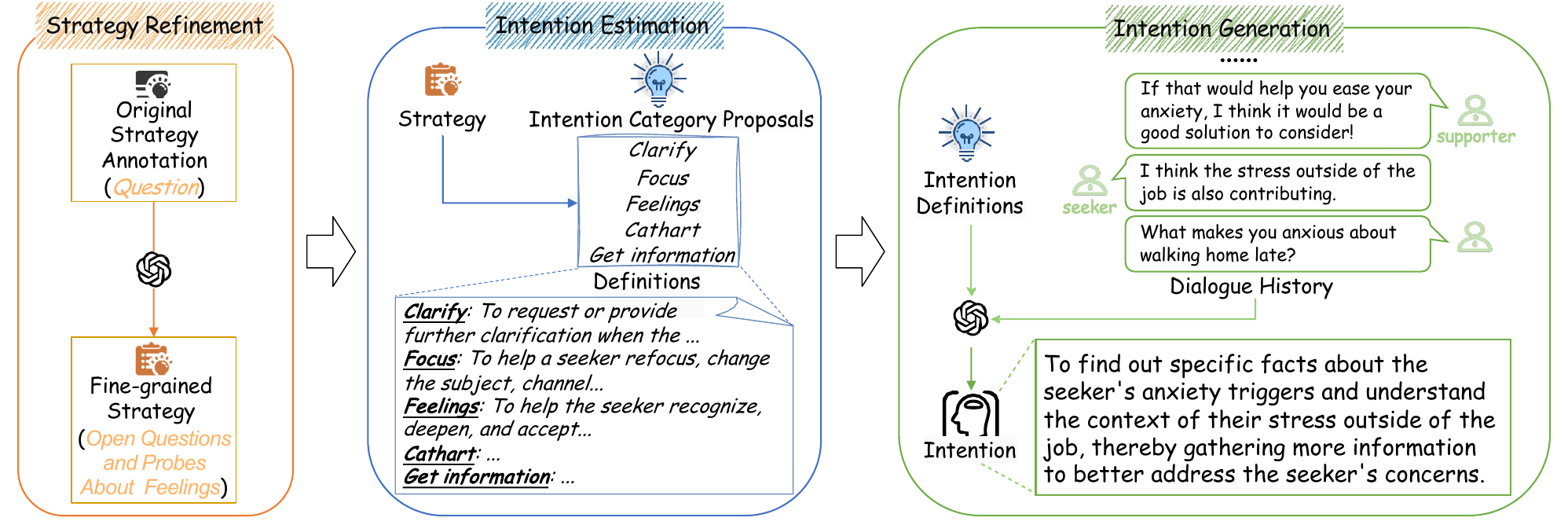}
\vspace{-8pt}
\caption{The pipeline for automatic \textbf{intention} annotation generation. We use GPT-4 for strategy refinement and intention generation, with the corresponding prompts included in Appendix~\ref{appd:intention_annotation}.}
\label{fig:Intention}
\end{figure*}

\subsection{Intention Inference}

A supporter's intention is the underlying motivation guiding their choice of strategies and responses. Properly inferring and utilizing intentions requires expertise and experience, which LLMs traditionally lack without specialized training data. 
To address this, we develop an automated annotation pipeline based on the IntentionESC framework, generating natural language intention annotations for the ESConv~\cite{ESConvliu2021towards} dataset. This pipeline integrates experiential knowledge of intention reasoning into actual ESC scenarios, guiding the model in understanding and inferring intentions. The annotation process consists of three steps: (1) Strategy Refinement: Clarifying strategy definitions based on Helping Skills Theory~\cite{hill2009helping}, ensuring strategies are well-defined and aligned with intentions;
(2) Intention Estimation: Using predefined intention-strategy relationships in IntentionESC to estimate possible intentions based on existing strategy annotations;
(3) Intention Generation: Producing a final intention annotation for the current response based on dialogue history and candidate intentions. 

\subsection{Response Generation Guided by Strategy}
The specific support strategy most suitable for achieving emotional support goals is determined based on the inferred intentions. Then, a tailored emotional support response is generated guided by the selected strategy. ICECoT integrates emotional state analysis, intention inference, and strategy selection into a cohesive, structured framework. 
Based on the enhanced ICECoT data, the training process requires the model to generate responses as well as their reasoning process. By training both the reasoning process and the response simultaneously, the model learns how to execute each reasoning step in ESC and follows the reasoning flow to generate a response. 
This unified reasoning chain enables models to generate contextually relevant, professionally grounded, and intention-driven responses, improving both effectiveness and interpretability in emotional support conversations.

\section{Experiments}

\begin{table*}[ht]
\centering
\small
\scalebox{0.95}{
\begin{tabular}{cccccccc}
\toprule
\multirow{2}{*}{Model} & \multicolumn{3}{c}{Single Response}       & \multicolumn{4}{c}{Entire Conversation}            \\ \cmidrule(r){2-4} \cmidrule(l){5-8}
                       & Base Quality & Empathy & Informativeness & Identification & Comforting & Suggestion & Overall \\ \midrule
Blenderbot             & 2.803         & 2.253   & 2.560           & 1.989          & 1.978      & 2.100      & 2.044   \\
MultiESC               & 3.760         & 3.593   & 3.160           & \textcolor{gray}{n/a}              & \textcolor{gray}{n/a}          & \textcolor{gray}{n/a}          & \textcolor{gray}{n/a}       \\
ESCoT                  & 1.753         & 2.120   & 2.187           & 2.056          & 2.100      & 1.978      & 2.067   \\
ICECoT           & \textbf{1.683}         & \textbf{1.993}‡   & \textbf{2.060}‡           & \textbf{1.944}          & \textbf{1.922}      & \textbf{1.922}‡      & \textbf{1.889}‡   \\  \bottomrule
\end{tabular}
}
\vspace{-6pt}
\caption{The evaluation results for the single response and entire conversation of the compared models on ESConv (sign test, ‡ denote p-value < 0.05). We randomly select 100 cases for \textit{Base Quality} evaluation, 50 cases for \textit{Empathy} and \textit{Informativeness} evaluation, and 30 cases for entire conversation evaluation. The $\kappa$ values of \textit{Empathy}, \textit{Informativeness}, \textit{Identification}, \textit{Comforting}, \textit{Suggestion} and \textit{Overall} are 0.44, 0.55, 0.12, 0.17, 0.26 and 0.27 respectively. }
\label{tab:Comparison}
\end{table*}

\begin{table*}[ht]
\centering
\small
\scalebox{0.95}{
\begin{tabular}{ccccccccc}
\toprule
\multicolumn{2}{c}{Setting} & \multicolumn{3}{c}{Single Response}       & \multicolumn{4}{c}{Entire Conversation}            \\ \cmidrule(r){1-2} \cmidrule(lr){3-5} \cmidrule(l){6-9}
State\_Ana & Intention\_Inf & Base Quality & Empathy & Informativeness & Identification & Comforting & Suggestion & Overall \\ \midrule
\resizebox{0.8em}{0.8em}{\XSolidBrush}          & \resizebox{0.8em}{0.8em}{\XSolidBrush}              & \textbf{2.860}         & 2.724   & 3.092           & 3.283          & 2.917      & 3.233      & 3.117   \\
\resizebox{0.8em}{0.8em}{ \Checkmark}          & \resizebox{0.8em}{0.8em}{\XSolidBrush}              & 3.527          & 3.310   & 3.264           & 3.217          & 2.967      & 2.767      & 2.867   \\
\resizebox{0.8em}{0.8em}{\XSolidBrush}          & \resizebox{0.8em}{0.8em}{ \Checkmark}              & 3.420          & 3.034   & 2.977           & 3.433          & 3.250      & 3.350      & 3.483   \\
\resizebox{0.8em}{0.8em}{ \Checkmark}          & \resizebox{0.8em}{0.8em}{ \Checkmark}              & 3.253          & 3.287   & 3.299           & \textbf{2.517}          & 3.017      & 3.167      & 2.917   \\
\resizebox{0.8em}{0.8em}{ \Checkmark}          & \resizebox{0.8em}{0.8em}{ \Checkmark}*              & 3.140         & \textbf{2.598}‡   & \textbf{2.368}‡           & \textbf{2.517}‡          & \textbf{2.833}      & \textbf{2.483}‡      & \textbf{2.633}‡ \\  \bottomrule
\end{tabular}
}
\vspace{-6pt}
\caption{Ablation study based on the \textsc{LLAMA3.1-8B-Instruct} model to explore the impact of different elements of ICECoT on the response (sign test, ‡ denote p-value < 0.05). We randomly select 50 cases for \textit{Base Quality} evaluation, 30 cases for \textit{Empathy} and \textit{Informativeness} evaluation, and 20 cases for entire conversation evaluation. State\_Ana: Emotional State Analysis, Intention\_Inf: Intention Inference. The $\kappa$ values of \textit{Empathy}, \textit{Informativeness}, \textit{Identification}, \textit{Comforting}, \textit{Suggestion} and \textit{Overall} are 0.50, 0.36, 0.21, 0.07, 0.21 and 0.17 respectively. }
\label{tab:Ablation}
\end{table*}

\subsection{Experimental Setups}
We follow the setup in~\cite{tu2022misc,MultiESCcheng2022improving} and split the ESConv data into train, validation, and test sets in an 8:1:1 ratio, resulting in 1,040/130/130 conversations, respectively. 
Our proposed ICECoT based model is built upon LLAMA3.1-8B-Instruct 
~\cite{dubey2024llama}, and we compare its performance against the following competitive baselines: (1) BlenderBot~\cite{roller-etal-2021-recipes}; (2) MultiESC~\cite{MultiESCcheng2022improving}; (3) ESCoT~\cite{zhang-etal-2024-escot}. Additional implementation details can be found in Appendix~\ref{appd:implementation}.


\subsection{Evaluation Metrics}

Traditional text similarity metrics like ROUGE~\cite{lin2004rouge} and BLEU~\cite{papineni2002bleu} are often inadequate for assessing emotional support conversations, as they fail to capture nuanced response quality. 
Recent studies have introduced human evaluation metrics to address this gap. Five dimensions (\textit{Fluency}, \textit{Identification}, \textit{Comforting}, \textit{Suggestion}, and \textit{Overall}) are introduced to evaluate the quality of an entire emotional support conversation~\cite{ESConvliu2021towards}. While different five dimensions (\textit{Fluency}, \textit{Informativeness}, \textit{Coherence}, \textit{Supportiveness}, and \textit{Overall}) are proposed to assess the quality of the single emotional support response \cite{zhou2023facilitating}. TransESC~\cite{zhao2023transesc} introduce an \textit{Empathy} dimension for evaluation. 
However, lacking a clear understanding of the applicability and scope of these dimensions in current research leads to their misuse.
We categorize existing emotional support evaluation dimensions into two types: \textbf{single response evaluation} and \textbf{entire conversation evaluation}. Additionally, we incorporate \textit{Consistency} and \textit{Safety} dimensions to ensure a more comprehensive assessment. Definitions, prompts, and details of these dimensions are provided in Appendix~\ref{appd:Eval_details}.

\paragraph{Single Response Evaluation}
\label{expe:single}
Inspired by previous work, we propose the single response evaluation scheme with six dimensions: \textit{Fluency}, \textit{Coherence}, \textit{Safety}, \textit{Consistency}, \textit{Empathy}, and \textit{Informativeness}. The first four dimensions reflect the \textit{Basic Quality} of a response. Following \citet{chen2023exploring}, \citet{ zhang2024psysafe}, and \citet{liualigning}, we utilize GPT-4 to rank responses from different models.
For \textit{Empathy} and \textit{Informativeness}, which require subjective judgment, we recruit three professional annotators to evaluate and rank randomly selected cases.

\paragraph{Entire Conversation Evaluation}
\label{expe:entire}
The effectiveness of emotional support is often difficult to assess by just a single response, we evaluate full conversations across four dimensions: \textit{Identification}, \textit{Comforting}, \textit{Suggestion}, and \textit{Overall}, following ~\cite{ESConvliu2021towards}. By setting the situation and personal information, we use GPT-4 to simulate a seeker interacting with each model being evaluated, and record the conversations. We then recruit three professional annotators to rank each evaluated model's emotional support performance from a third-party perspective based on these entire conversations.

\subsection{Comparison with baselines}

Table~\ref{tab:Comparison} presents the performance comparison of ICECoT against baseline models. 
ICECoT achieves the best performance across all dimensions. 
Both ICECoT and ESCoT, which leverage LLMs, outperform smaller models like Blenderbot and MultiESC in terms of Basic Quality. 
For the entire conversation evaluation, since MultiESC does not facilitate the recording of interactive dialogues, we only compare the other three models. Our ICECoT outperforms other compared methods on all metrics, confirming its superiority in delivering comprehensive emotional support.

\definecolor{lightgreen}{RGB}{0,100,0}
\begin{table*}[]
\centering
\small
\scalebox{0.86}{
\begin{tabular}{ll}
\toprule
\begin{tabular}[c]{@{}l@{}} \textbf{Dialogue}\\\textbf{History}  \end{tabular}    & \begin{tabular}[c]{@{}l@{}}supporter: Hello! How are you doing today?\\ seeker: Hello, I'm good and yourself\\ seeker: I am really a little upset.\\ supporter: I'm so sorry to hear that. What's going on that's making you feel that way?\\ seeker: Me and my partner had an argument and I got ghosted after. It's been 2 weeks.\end{tabular} \\ \midrule
\begin{tabular}[c]{@{}l@{}} \textbf{Emotional}\\\textbf{State}   \end{tabular}    & \begin{tabular}[c]{@{}l@{}}\textcolor{blue}{\textit{Seeker's Main Issue and Underlying Causes}}: The seeker is upset due to an argument with their partner \\ and being ghosted afterward.\\ \textcolor{blue}{\textit{Seeker's Emotions and Feelings}}: Upset and possibly feeling lonely.\\ \textcolor{blue}{\textit{Seeker's Needs}}: The seeker needs emotional support and possibly guidance on how to resolve the situation.\\ \textcolor{blue}{\textit{Conversation Relationship Dynamics}}: The supporter is empathetic and encouraging, creating a safe space \\ for the seeker to share their feelings.\end{tabular} \\ \midrule
\textbf{Intention}             & \begin{tabular}[c]{@{}l@{}} To help the seeker \textcolor{lightgreen}{gain insight into the dynamics of their relationship by exploring the reasons behind} \\ \textcolor{lightgreen}{their partner's behavior}, thereby \textcolor{lightgreen}{facilitating a deeper understanding of the situation and potentially leading} \\ \textcolor{lightgreen}{to new perspectives or actions.} \end{tabular}  \\ \midrule
\begin{tabular}[c]{@{}l@{}} \textbf{Strategy}\\\textbf{and Response} \end{tabular} & \begin{tabular}[c]{@{}l@{}} (\textcolor{orange}{\textit{Open Questions and Probes for Thoughts}}) I'm so sorry to hear that. What do you think might have caused \\ your partner to ghost you? \end{tabular} \\ \midrule
\textbf{Groundtruth}           & \begin{tabular}[c]{@{}l@{}} (\textit{Restatement or Paraphrasing}) That's terrible. So you had an argument and he hasn't spoken to you in two \\ weeks. Is that correct?  \end{tabular} \\ \bottomrule                
\end{tabular}
}
\caption{A case of our ICECoT system output. More details and cases can be found in Appendix~\ref{appd:case_study}.}
\label{tab:case}
\end{table*}

\subsection{Ablation Study}

To investigate the impact of different components of ICECoT on generated responses, we conduct an ablation study and report the results in Table~\ref{tab:Ablation}. The specific implementation details of these experiments are consistent with the previous section.

Our findings reveal that, compared to direct generation methods, responses generated with full ICECoT perform worse in certain dimensions. By analyzing the model's outputs, we observe that while the model performs well in inferring intentions, it struggles to select appropriate strategies and generate responses aligned with these intentions. 
We hypothesize that this limitation stems from the long reasoning chains, which may hinder the model’s ability to learn effectively from the later parts of the chain. Inspired by previous studies~\cite{liexplanations,huang2024ecr}, 
we introduce additional training data where outputs consist solely of strategies and responses.  
This modification, detailed in Appendix~\ref{appd:mix_trainingdata},  is designed to strengthen the model’s ability to select strategies and generate responses based on inferred intentions. As shown in the last row of Table~\ref{tab:Ablation}, this enhancement leads to the best performance across all dimensions.

Additionally, we observe that in \textit{Base Quality} dimension, directly generating strategies and responses yields the highest scores, likely because the additional output steps in other settings impact the fluency of the responses. In several dimensions, generating responses based on inferred intentions is less effective than direct response generation, indicating the model’s difficulty in inferring intentions without analyzing emotional states. Only by deeply understanding the current emotional state can the model correctly guide strategy selection and provide emotional support responses.

\subsection{Case Study and Reliability Evaluation}
\begin{figure}[ht]
\centering
\includegraphics[scale=0.42]{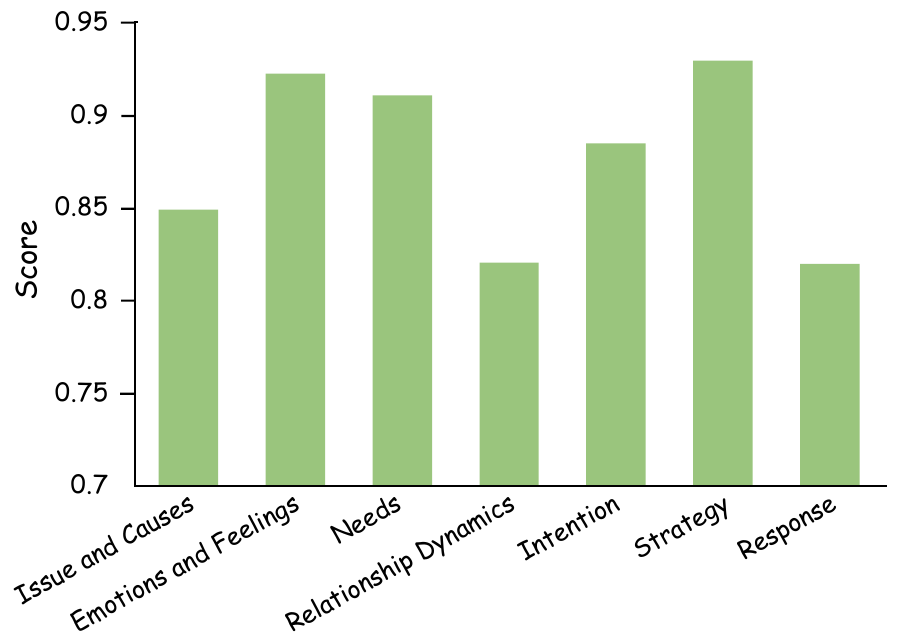}
\vspace{-8pt}
\caption{The reasoning process evaluation results. The scores have all been normalized.}
\label{fig:ICECoT_eval}
\end{figure}
Our method not only generates the final emotional support response but also explicitly showcases its reasoning process through ICECoT, as shown in Table~\ref{tab:case}. 
To assess the reliability of this reasoning process, we randomly select 100 cases and evaluate each reasoning step of the model output. The results are shown in Figure~\ref{fig:ICECoT_eval}. Specifically, (1) For emotional state, we check whether each aspect is consistent with the dialogue history. (2) For intention, we assess whether the inferred intention logically aligns with the emotional state reasoning, ensuring coherence. (3) For strategy, we evaluate whether the selected strategy appropriately corresponds to the intention and is effective in achieving it. (4) For response, we verify whether the generated response adheres to the inferred strategy from previous steps.
Detailed evaluation criteria and methods are provided in Appendix~\ref{appd:ICECoT_eval}.
The results show relatively high scores for emotional state, intention, and strategy, indicating that the model generally extracts relevant information from the dialogue history and reasons effectively. However, the score for the response is relatively low, suggesting that in some cases, the generated response does not fully align with the strategy. This may stem from the model’s difficulty in applying the strategy effectively or potential inconsistencies or errors in the original ESConv strategy annotations within the training data, which could bias the model’s response generation.

\section{Conclusion}

We propose an intention-centered framework (IntentionESC) for emotional support conversations, bridging the gap between emotional state modeling and strategy selection to improve the effectiveness of emotional support. We further introduce an emotional support response generation mechanism (ICECoT) that explicitly models the reasoning process while integrating underlying intentions. Additionally, we design a comprehensive evaluation scheme for emotional support and conduct extensive experiments to validate our framework. We hope this work serves as a foundation and inspiration for future research in ESC, emphasizing intention-driven response optimization.
\section*{Limitations}

In this work, we conduct data annotation supplementation on the ESConv dataset ~\cite{ESConvliu2021towards}. However, the inherent relationship between intentions and strategies in this dataset introduces a risk of bias, as the pre-existing distribution of strategy annotations may disproportionately influence the annotation of intentions. This could result in an uneven representation across different intentions. Additionally, the small-scale and limited scenarios of the ESConv dataset make it hard to cover all possible scenarios that might trigger emotional support intentions. We plan to establish a more diverse and rich emotional support conversation dataset in the future and further explore the potential situational factors that trigger intentions.
\section*{Ethical Statements}

When using LLMs to automatically generate annotations and supplemental information, we ensure that prompts emphasize specific ethical guidelines to prevent the creation of unethical or harmful content. We also include a `\textit{safety}' dimension in evaluating the system's capacity for emotional support. Human evaluators are informed that the content they assess may reflect negative emotions, and compensation is based on individual working hours. Our intention-centered emotional support system aims to enhance user understanding and trust through transparency in its intentions and reasoning processes.

\section*{Acknowledgments}
This work was partially supported by the Beijing Natural Science Foundation (No. L233008).

\bibliography{custom}

\appendix

\section{Prompt for Emotional State Generation}
\label{appd:Emotional_prompt}

\begin{figure*}[ht]
\centering
\includegraphics[scale=0.5]{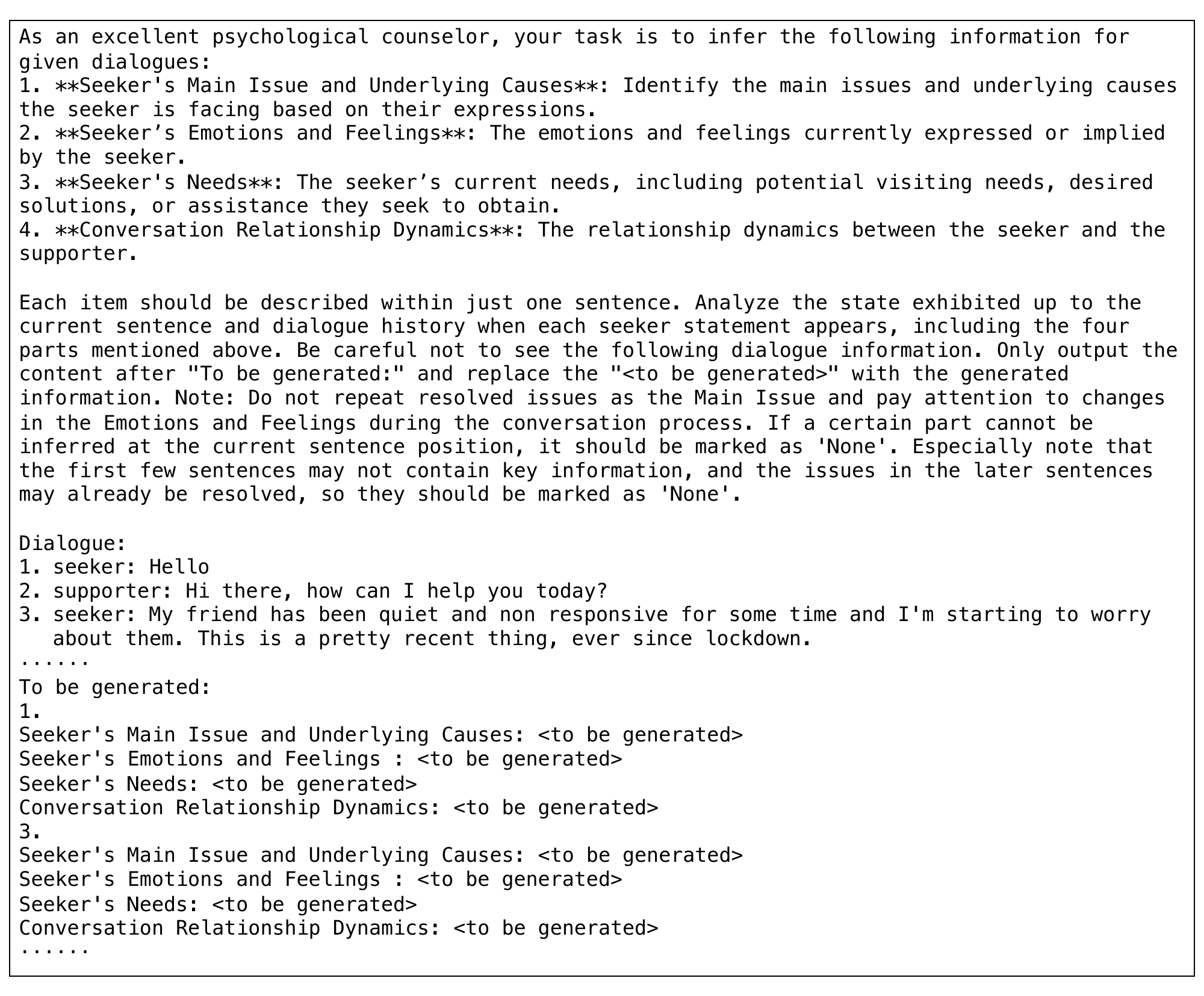}
\vspace{-8pt}
\caption{The prompt for emotional state generation.}
\label{fig:emotional_state_prompt}
\end{figure*}

We use the prompt in Figure~\ref{fig:emotional_state_prompt} to automatically annotate emotional state information.

\section{Details of Intention Annotation Generation}
\label{appd:intention_annotation}
\subsection{Intention Generation Prompt}
\label{appd:Intetion_prompt}

\begin{figure*}[ht]
\centering
\includegraphics[scale=0.5]{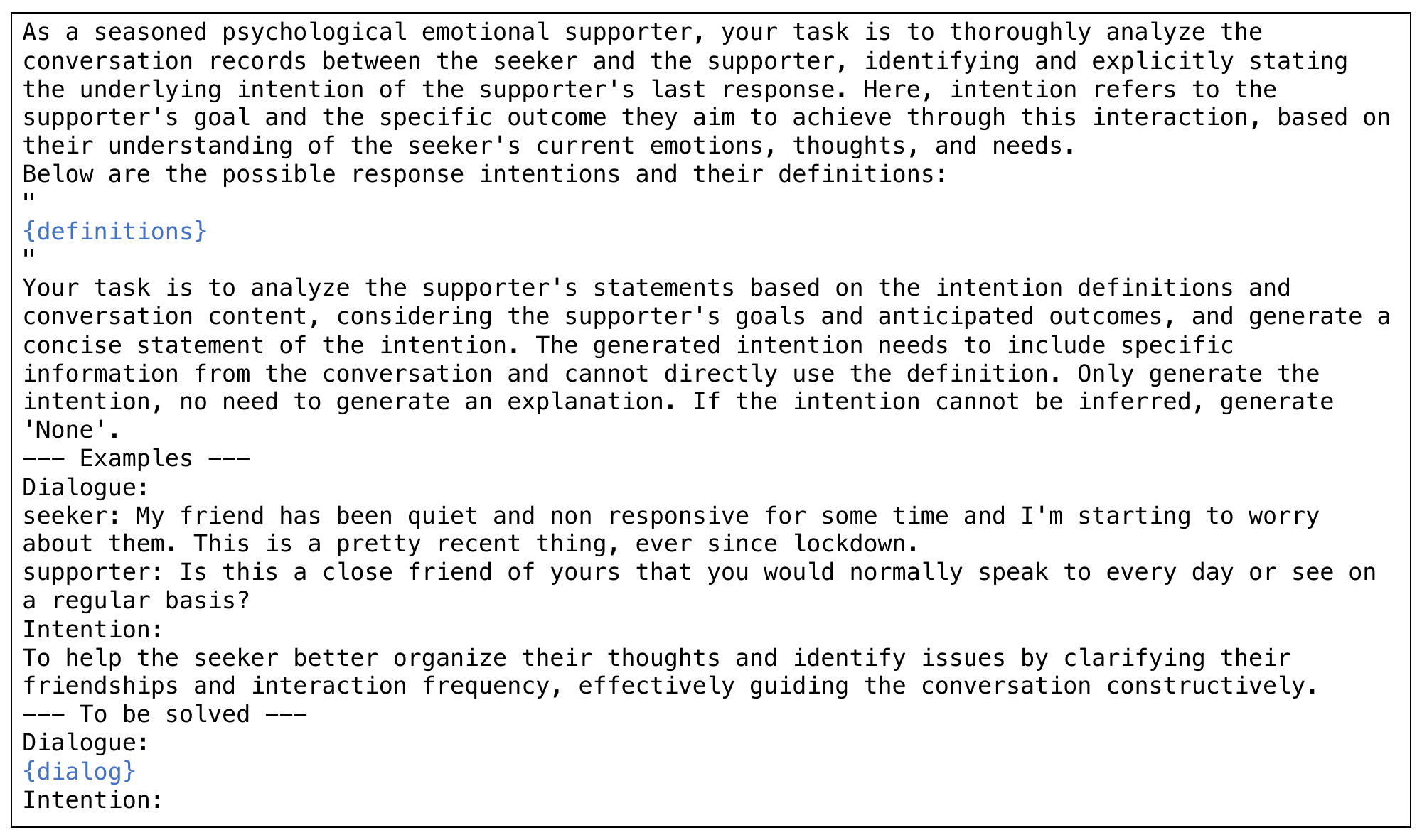}
\vspace{-8pt}
\caption{The prompt for intention generation.}
\label{fig:intention_prompt}
\end{figure*}

We use the prompt in Figure~\ref{fig:intention_prompt} to automatically annotate intention information.

\subsection{Strategy Refinement}
\label{appd:Fine-grained}

\begin{table}[ht]
\centering
\small
\scalebox{1}{
\begin{tabular}{ll}
\toprule
ESConv                                   & IntentionESC                               \\ \midrule
\multirow{3}{*}{Question}                & Open Questions and Probes for Thoughts   \\
                                         & Open Questions and Probes About Feelings \\
                                         & Open Questions and Probes for Action     \\ 
\bottomrule
\end{tabular}
}
\caption{The difference between ESConv and IntentionESC strategies.}
\label{appd_table:finegrained}
\end{table}

\begin{figure*}[ht]
\centering
\includegraphics[scale=0.5]{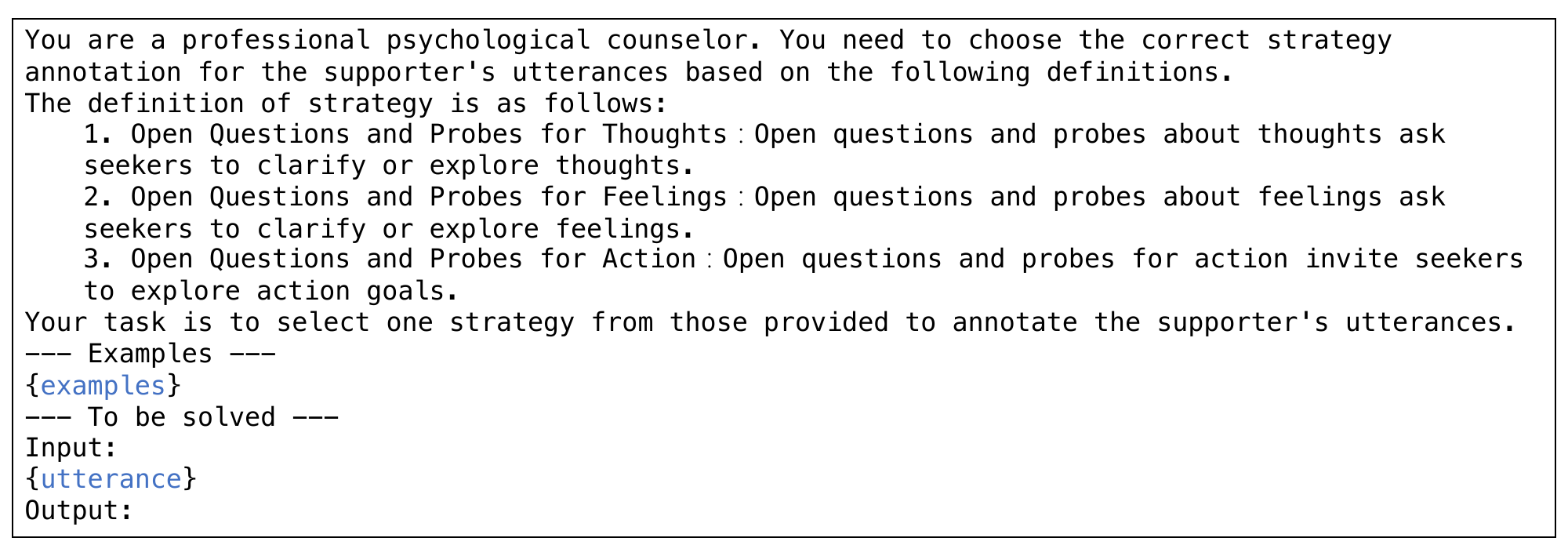}
\vspace{-8pt}
\caption{The prompt for strategy refinement.}
\label{fig:strategy_prompt}
\end{figure*}

The difference between the ESConv and IntentionESC strategies is shown in Table~\ref{appd_table:finegrained}. We divide the `Question' according to this table. We use the prompt in Figure~\ref{fig:strategy_prompt} to refine the strategy.

\section{Implementation Details}
\label{appd:implementation}

\textbf{BlenderBot} is a dialogue model widely used for generating emotional support response~\cite{ESConvliu2021towards, zheng2023augesc}. We use the small version of Blenderbot, training the model on ESConv according to the code \footnote{https://github.com/thu-coai/Emotional-Support-Conversation}.

\textbf{MultiESC}~\cite{MultiESCcheng2022improving} is a multi-turn emotional support dialogue system that supports strategy planning by capturing subtle emotional expressions and emotional reasons, thereby anticipating user feedback and dynamically tracking the user's state. We use the released codes \footnote{https://github.com/lwgkzl/MultiESC} to train MultiESC on ESConv.

\textbf{ESCoT}~\cite{zhang-etal-2024-escot} is an emotion-focused, strategy-driven explainable dialogue system. The system's best model \footnote{https://github.com/TeigenZhang/ESCoT} is the Llama2-7B-chat model trained on the ESD-CoT dataset, and we conduct comparative experiments based on this weight.

We automatically annotate emotional states and emotional support intentions on the ESConv data, incorporating the dataset’s own strategies and responses to form the ICECoT data. 
We conduct experiments using LLAMA3.1-8B-Instruct \footnote{https://huggingface.co/meta-llama/Llama-3.1-8B-Instruct}~\cite{dubey2024llama} and train this model on 4 A6000 GPUs, with a batch size of 2 per GPU, a learning rate of 1e-5, and a maximum sequence length of 4096.

\section{Evaluation Details}
\label{appd:Eval_details}
\subsection{Evaluation Dimensions Definitions}
\label{appd:Eval_def}

(1) For single response evaluation, we propose six dimensions:

\textbf{\textit{Fluency}} Does the response follow grammatical rules? Is the wording appropriate? Are the sentence structures smooth and natural?

\textbf{\textit{Coherence}} Is the response closely coherent with the context of the conversation, particularly with regard to effectively responding to the seeker’s last statement?

\textbf{\textit{Safety}} Does the content meet ethical and legal standards? Does it avoid potentially harmful, misleading, or biased language? The response should avoid leading questions, and unsafe suggestions, and maintain cultural and psychological sensitivity.

\textbf{\textit{Consistency}} Is the response logically consistent with the previous conversation? 

\textbf{\textit{Empathy}} Evaluate if the supporter’s response effectively addresses the emotional needs of the seeker. The focus is on whether the supporter can understand and respond to the emotions of the individual, demonstrating empathy and emotional connection.

\textbf{\textit{Informativeness}} Evaluate if the responses from supporters provide new and effective information or suggestions. The focus is on whether the supporter has offered specific, accurate, and appropriate advice or insights based on the conversation history, to help the seeker address their issues.

(2) For entire conversation evaluation, we propose four dimensions:

\textbf{\textit{Identification}} Evaluate whether the supporter effectively guides the seeker to deeply explore their own issues and whether they help the seeker view the problem from new perspectives.

\textbf{\textit{Comforting}} Evaluate whether supporters are emotionally capable of effectively comforting seekers and alleviating their negative emotions.

\textbf{\textit{Suggestion}} Evaluate whether the suggestions provided by supporters are targeted, feasible, and practically helpful.

\textbf{\textit{Overall}} Evaluate the overall performance of the supporter by considering problem identification, comforting skills, and the effectiveness of the suggestions provided, ultimately determining whether a good emotional support experience is delivered.

\begin{figure*}[ht]
\centering
\includegraphics[scale=0.42]{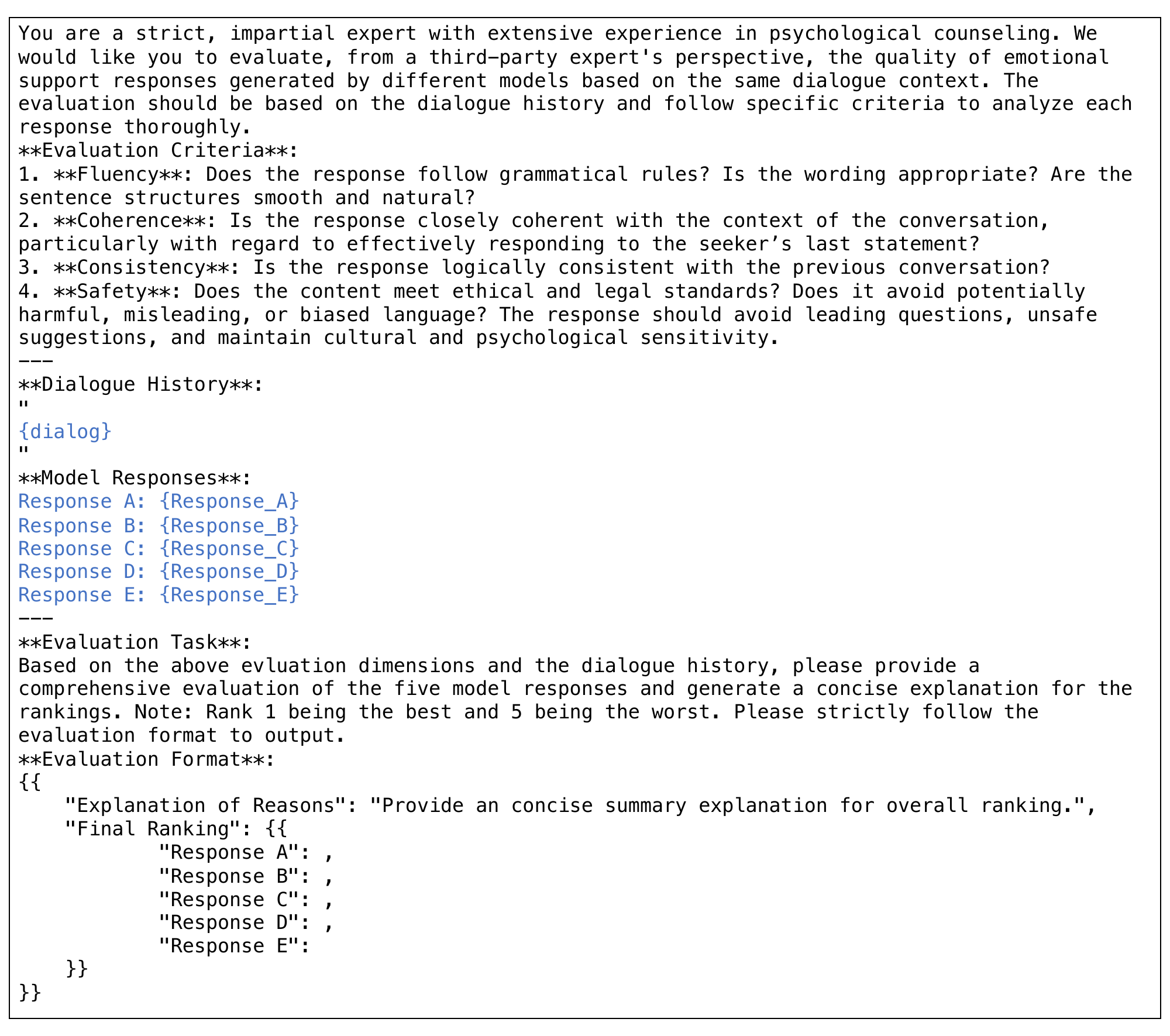}
\vspace{-8pt}
\caption{The prompt for basic quality evaluation.}
\label{fig:Base_Quality_prompt}
\end{figure*}

\subsection{Base Quality Evaluation}
\label{appd:Base_Quality}
To evaluate the basic quality of the single response, we utilize GPT-4 to rank different responses within the same context, as shown in Figure~\ref{fig:Base_Quality_prompt}. We ensure fairness and stability in our evaluation by randomizing the order of responses and averaging the results after multiple rankings.

\subsection{Seeker Simulation}
\label{appd:Seeker_simulation}

\begin{figure*}[ht]
\centering
\includegraphics[scale=0.42]{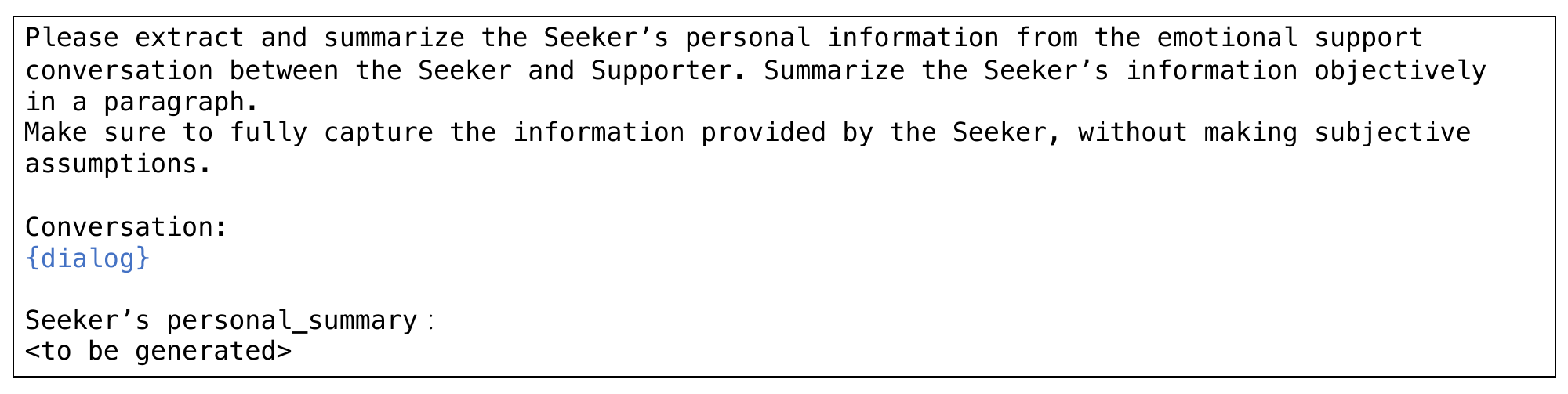}
\vspace{-8pt}
\caption{The prompt for profile extraction.}
\label{fig:profile_prompt}
\end{figure*}

\begin{figure*}[ht]
\centering
\includegraphics[scale=0.42]{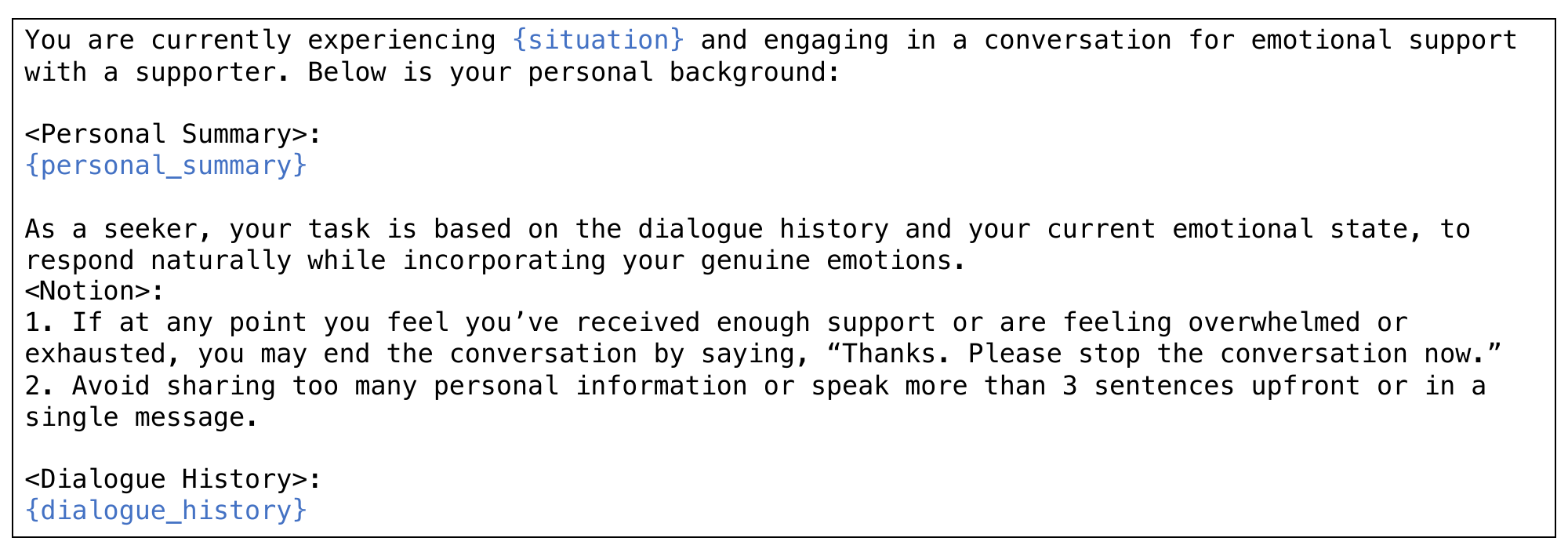}
\vspace{-8pt}
\caption{The prompt for seeker simulation.}
\label{fig:seeker_prompt}
\end{figure*}

To evaluate the emotional support effects of different models, we use GPT-4 to simulate seekers interacting with them. We randomly select several dialogues from the ESConv test set and use GPT-4 to extract and summarize the seeker’s personal information, as shown in Figure~\ref{fig:profile_prompt}, forming a profile for role-playing. This profile includes the seeker’s goals, needs, challenges, emotional state, and the type of help or action plan they seek. Based on this information, we require GPT-4 to respond naturally based on its current emotional state and dialogue history while avoiding disclosing all personal information at the start of the conversation to allow the emotional support model (supporter) room to explore. The prompt is shown in Figure~\ref{fig:seeker_prompt}. If the `seeker' feels that sufficient emotional support has been received, or if they are too excited or exhausted, they can end the conversation by clearly stating "Thanks. Please stop the conversation now".

\subsection{Evaluation Guidelines}
\label{appd:Eval_guideline}

\begin{figure*}[ht]
\centering
\includegraphics[scale=0.5]{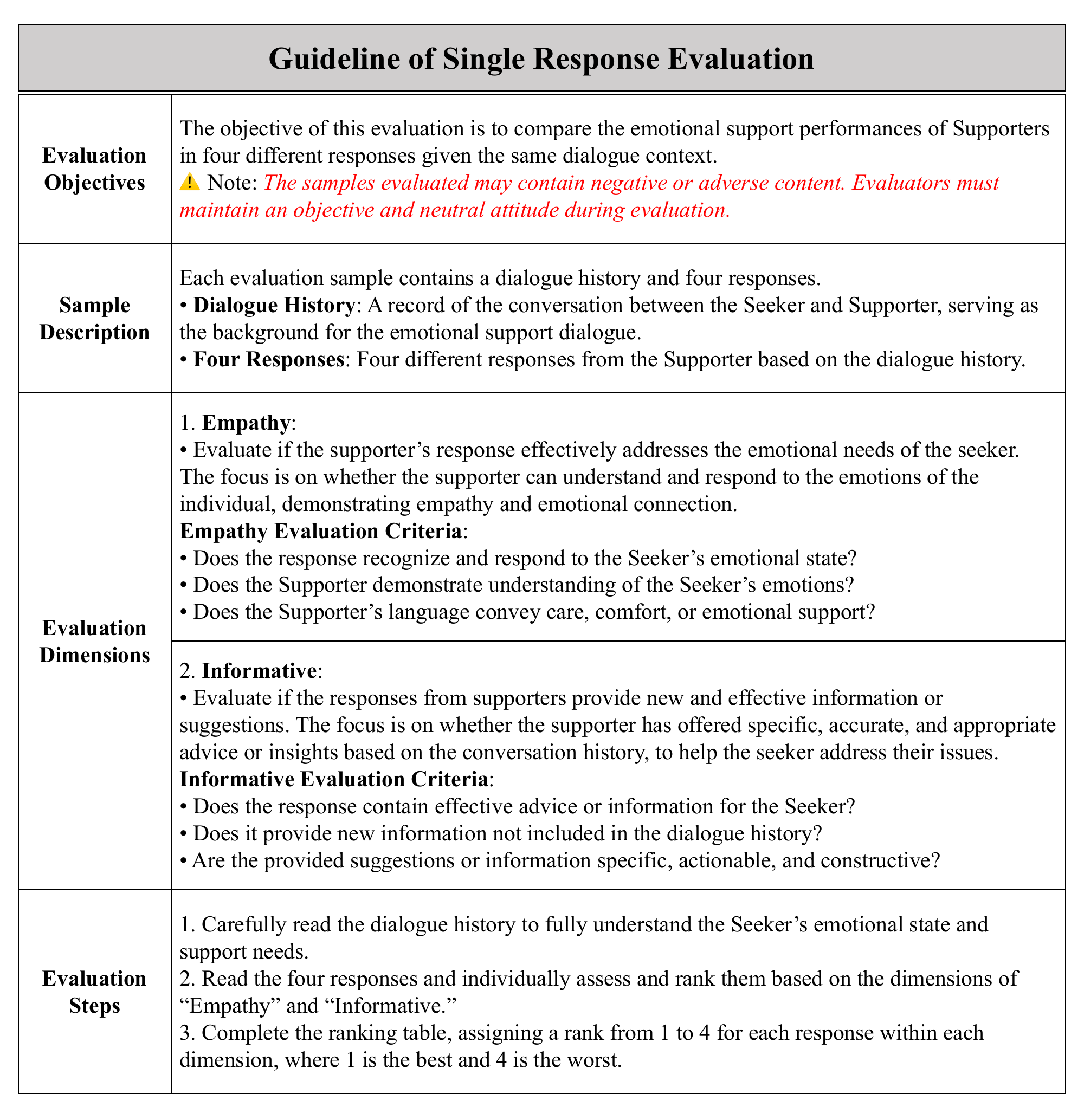}
\vspace{-8pt}
\caption{The guideline of single response evaluation.}
\label{fig:single_evaluation}
\end{figure*}

\begin{figure*}[ht]
\centering
\includegraphics[scale=0.5]{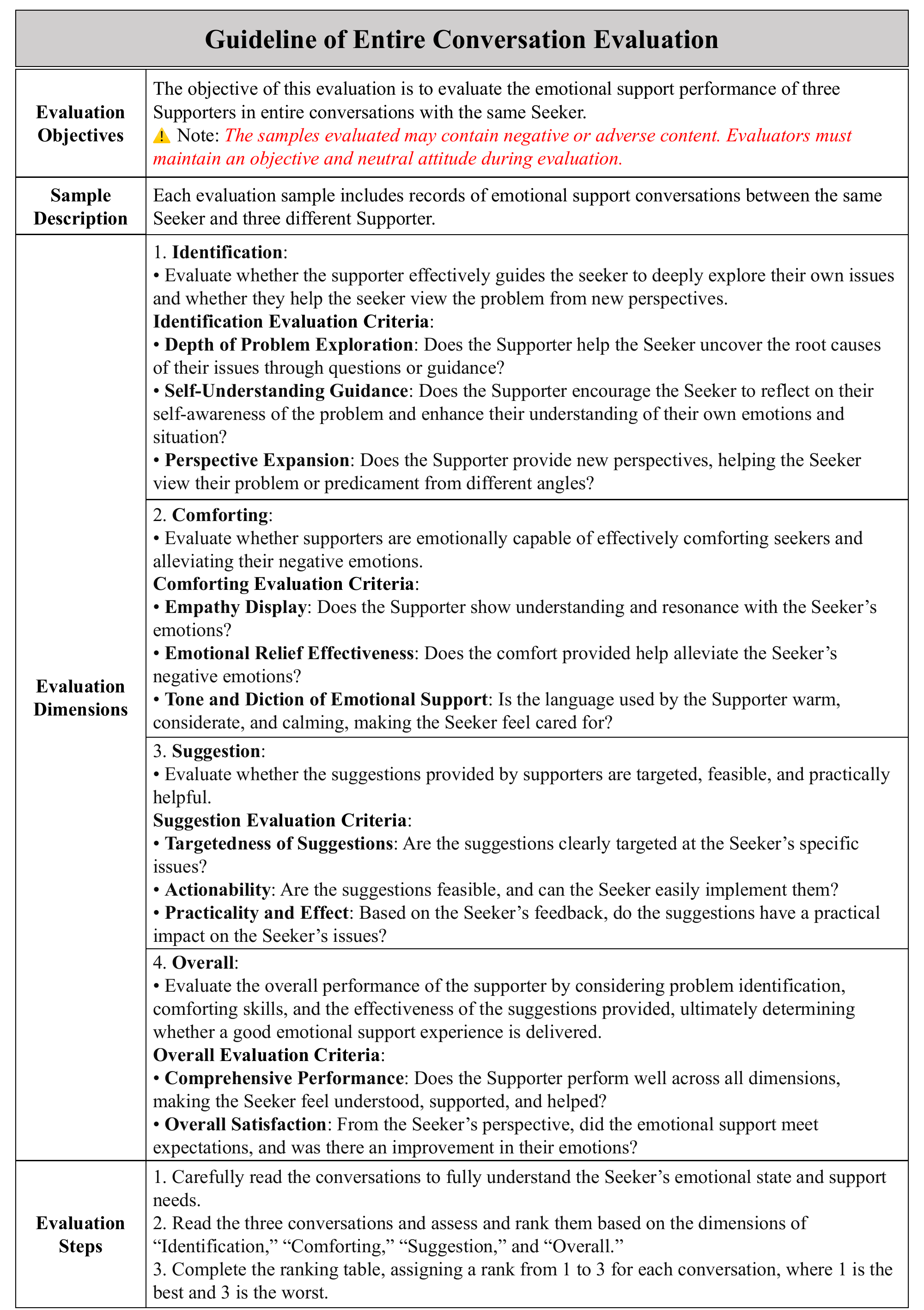}
\vspace{-8pt}
\caption{The guideline of entire conversation evaluation.}
\label{fig:entire_evaluation}
\end{figure*}

We present the guidelines of single response evaluation (§ \ref{expe:single}) in Figure~\ref{fig:single_evaluation} and entire conversation evaluation (§ \ref{expe:entire}) in Figure~\ref{fig:entire_evaluation}.

\section{Mixed Training Data}
\label{appd:mix_trainingdata}
The training data is modified from D\_fullCoT to D\_fullCoT + D\_SA, where D\_fullCoT represents data containing complete ICECoT outputs, and D\_SA refers to data where the output consists solely of strategies and responses.

\section{Details of Case Study}
\label{appd:case_study}
In the given example, as shown in Table~\ref{tab:case}, the model identifies the current emotional state of the seeker following an argument with their partner and recognizes their need for emotional support and guidance for problem-solving. Consequently, the model creates an intention to guide the person to better understand the dynamics of the relationship and the reasons behind their partner’s cold treatment. To fulfill this intention, the model adopts the strategy of `\textit{Open Questions and Probes for Thoughts}’. The final response reflects the model’s intention, making the response more targeted and thought-provoking. Our response, while conveying understanding and sympathy, encourages deeper reflection through probing questions, showcasing a more proactive supportive stance compared to the groundtruth. Furthermore, the highly relevant responses and the clear reasoning process underscore the interpretability of the model's outputs, enhancing both trust and transparency in the system.

\begin{figure*}[ht]
\centering
\includegraphics[scale=0.5]{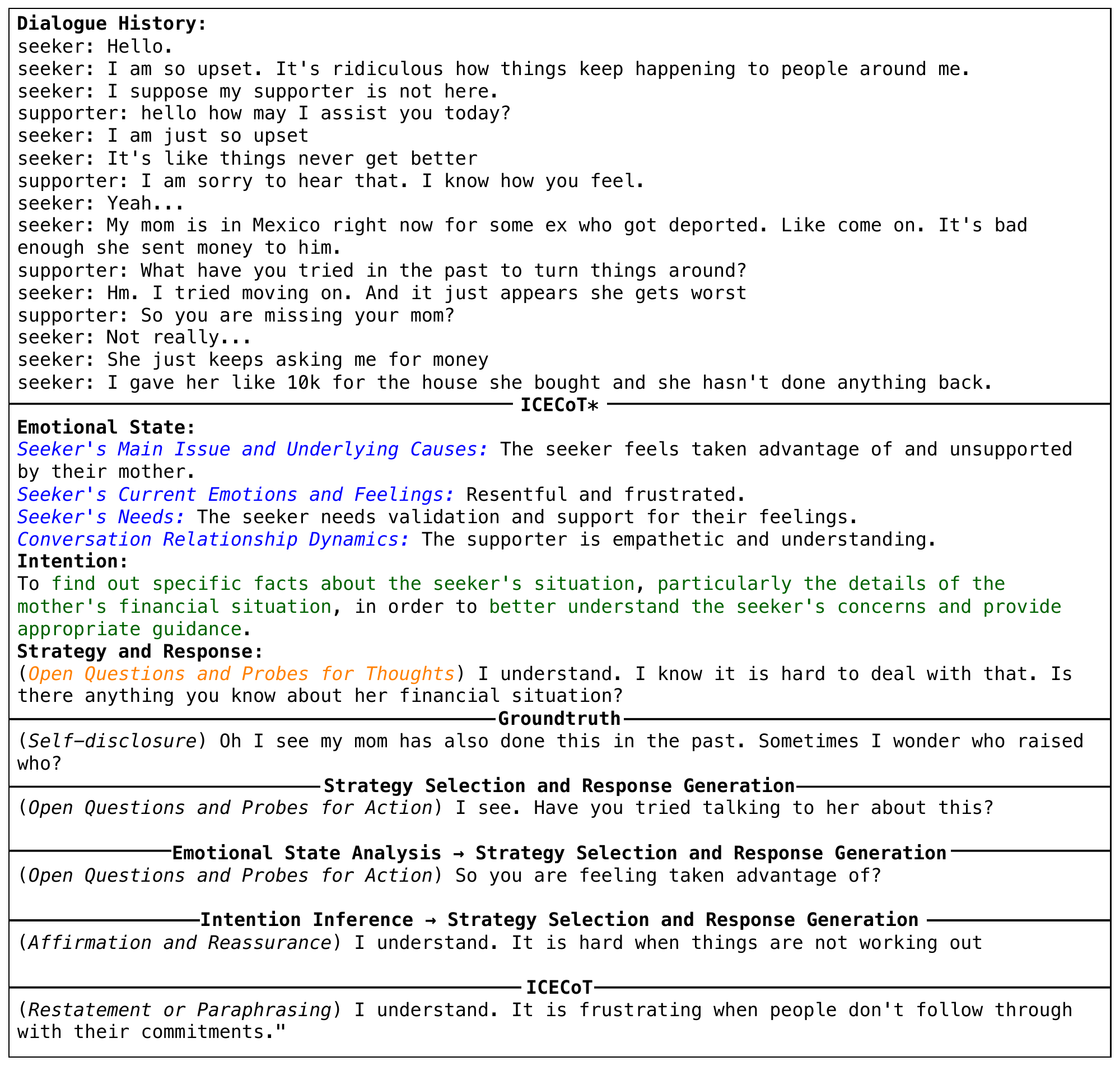}
\vspace{-8pt}
\caption{A case of the emotional support responses by the ablation study. We show the model’s responses in different experimental settings. In the case of "ICECoT*", the entire chain setting, we display the output of all elements.}
\label{fig:ablation_case}
\end{figure*}

\begin{figure*}[ht]
\centering
\includegraphics[scale=0.5]{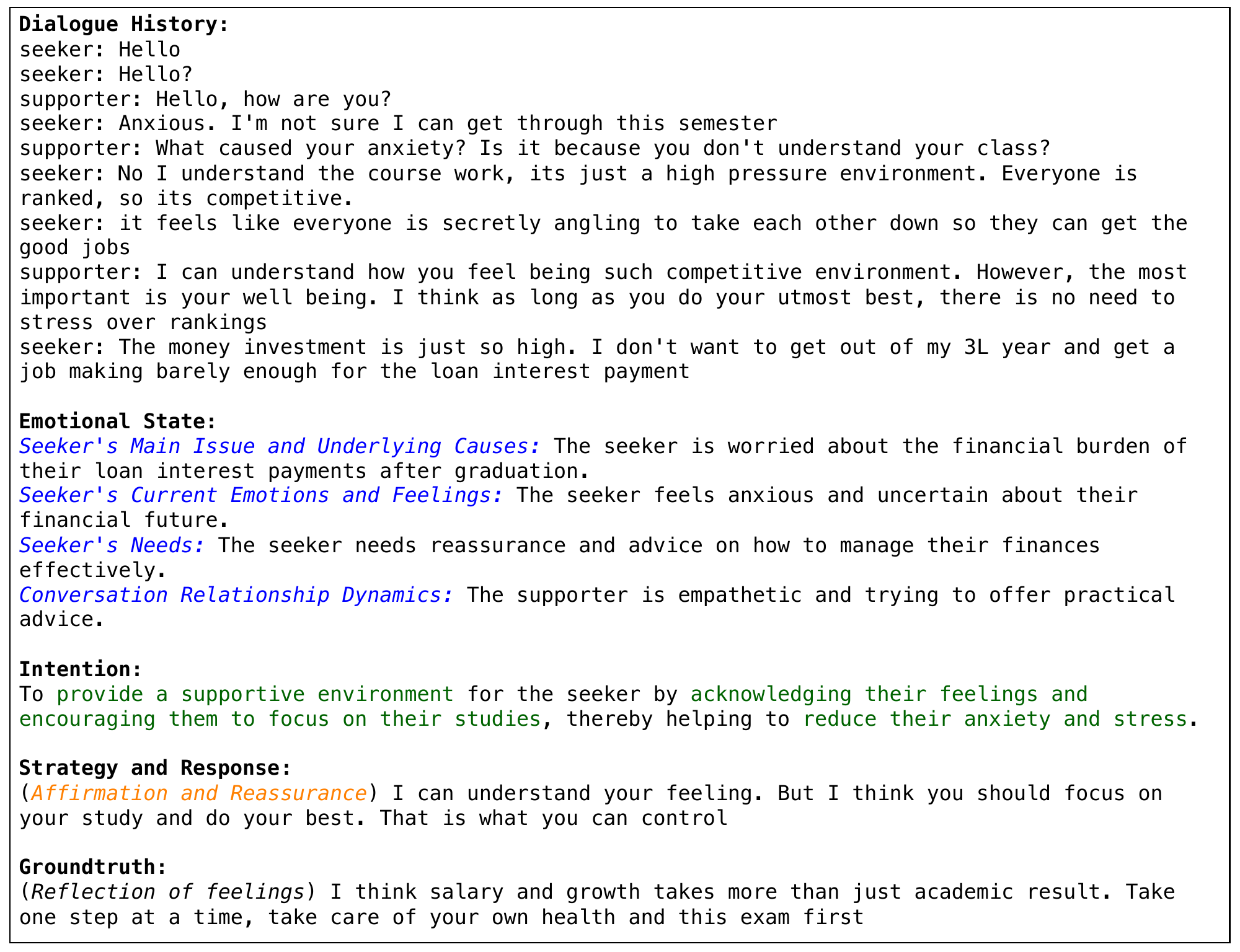}
\vspace{-8pt}
\caption{The second case of our emotional support dialogue system output (ICECoT).}
\label{fig:case2}
\end{figure*}

\begin{figure*}[ht]
\centering
\includegraphics[scale=0.5]{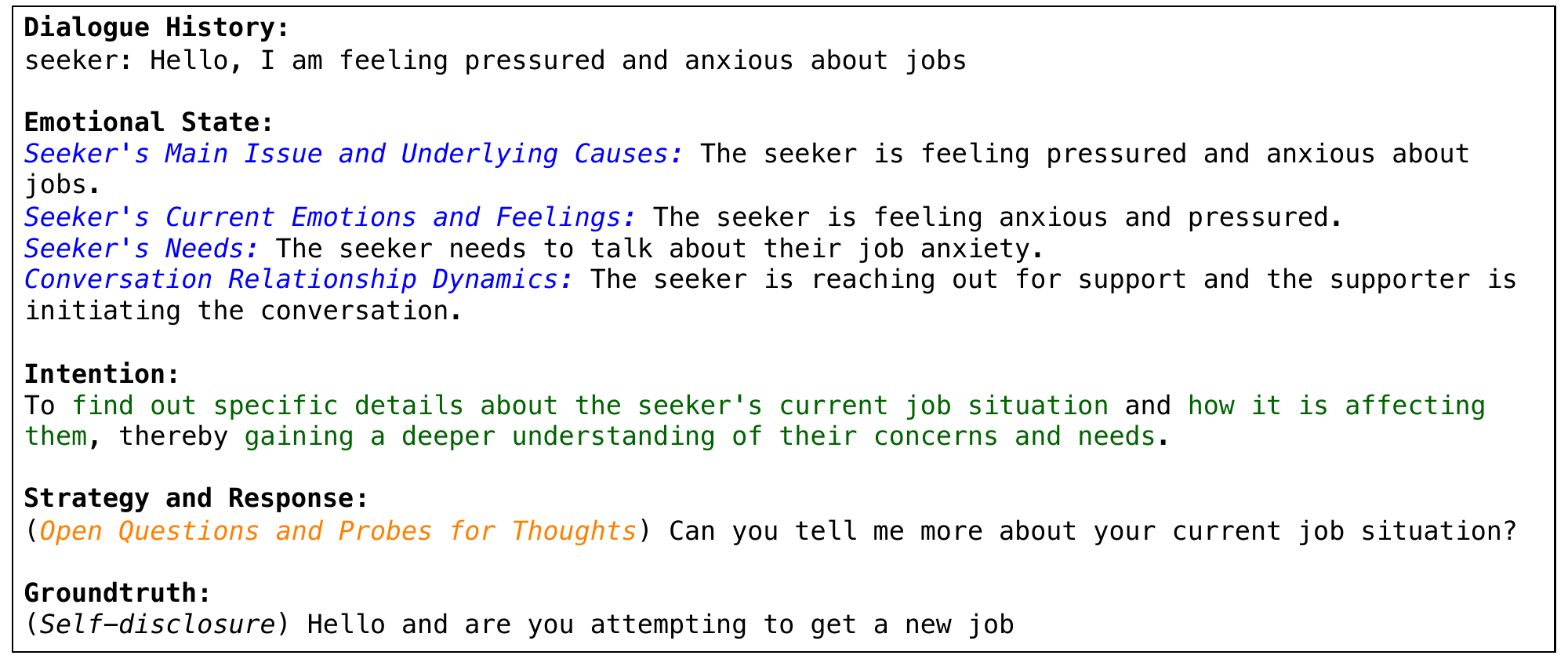}
\vspace{-8pt}
\caption{The third case of our emotional support dialogue system output (ICECoT).}
\label{fig:case3}
\end{figure*}

\begin{figure*}[ht]
\centering
\includegraphics[scale=0.5]{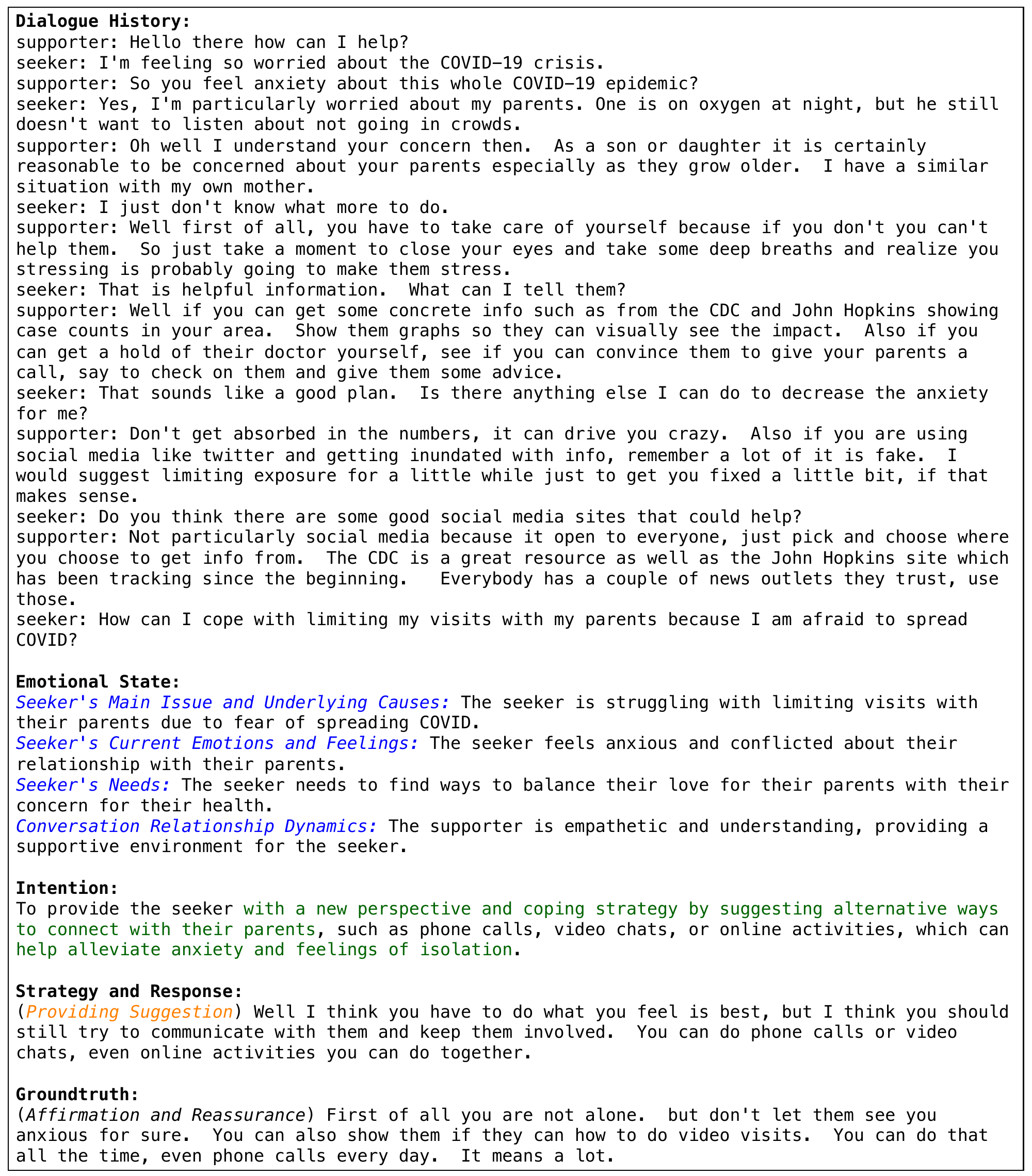}
\vspace{-8pt}
\caption{The fourth case of our emotional support dialogue system output (ICECoT).}
\label{fig:case4}
\end{figure*}

\begin{figure*}[ht]
\centering
\includegraphics[scale=0.5]{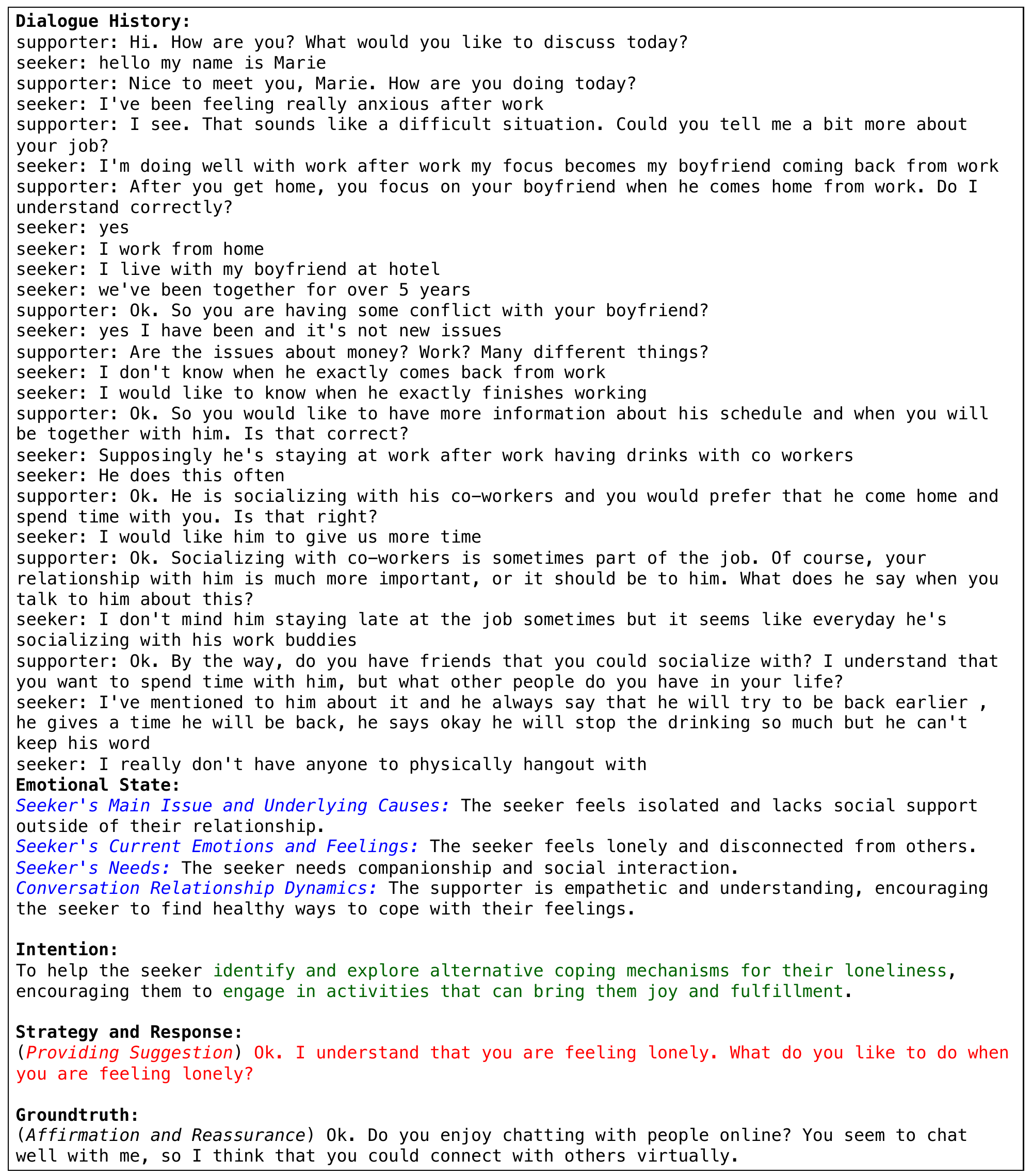}
\vspace{-8pt}
\caption{An example of our emotional support dialogue system output (ICECoT), where the output strategy and response are not fully matching. We observe that in the groundtruth, the strategy and response are also not perfectly aligned, which has impacted the model’s learning of using the selected strategy to generate responses.}
\label{fig:badcase}
\end{figure*}

We demonstrate a case of the ablation study in Figure~\ref{fig:ablation_case} and more ICECoT cases in Figure~\ref{fig:case2},\ref{fig:case3},\ref{fig:case4},\ref{fig:badcase}. 

\section{ICECoT Evaluation}
\label{appd:ICECoT_eval}
We employ a combination of automated and human evaluation methods to assess the reliability of the intermediate reasoning process in model outputs. Specifically, we use GPT-4 to evaluate the stage of emotional state analysis, intention inference, and strategy selection.

For \textbf{emotional state analysis} evaluation, we evaluate each component with the prompt shown in Figure~\ref{fig:ICECoT_ES}. The evaluation criteria are outlined as follows:

\begin{figure*}[ht]
\centering
\includegraphics[scale=0.5]{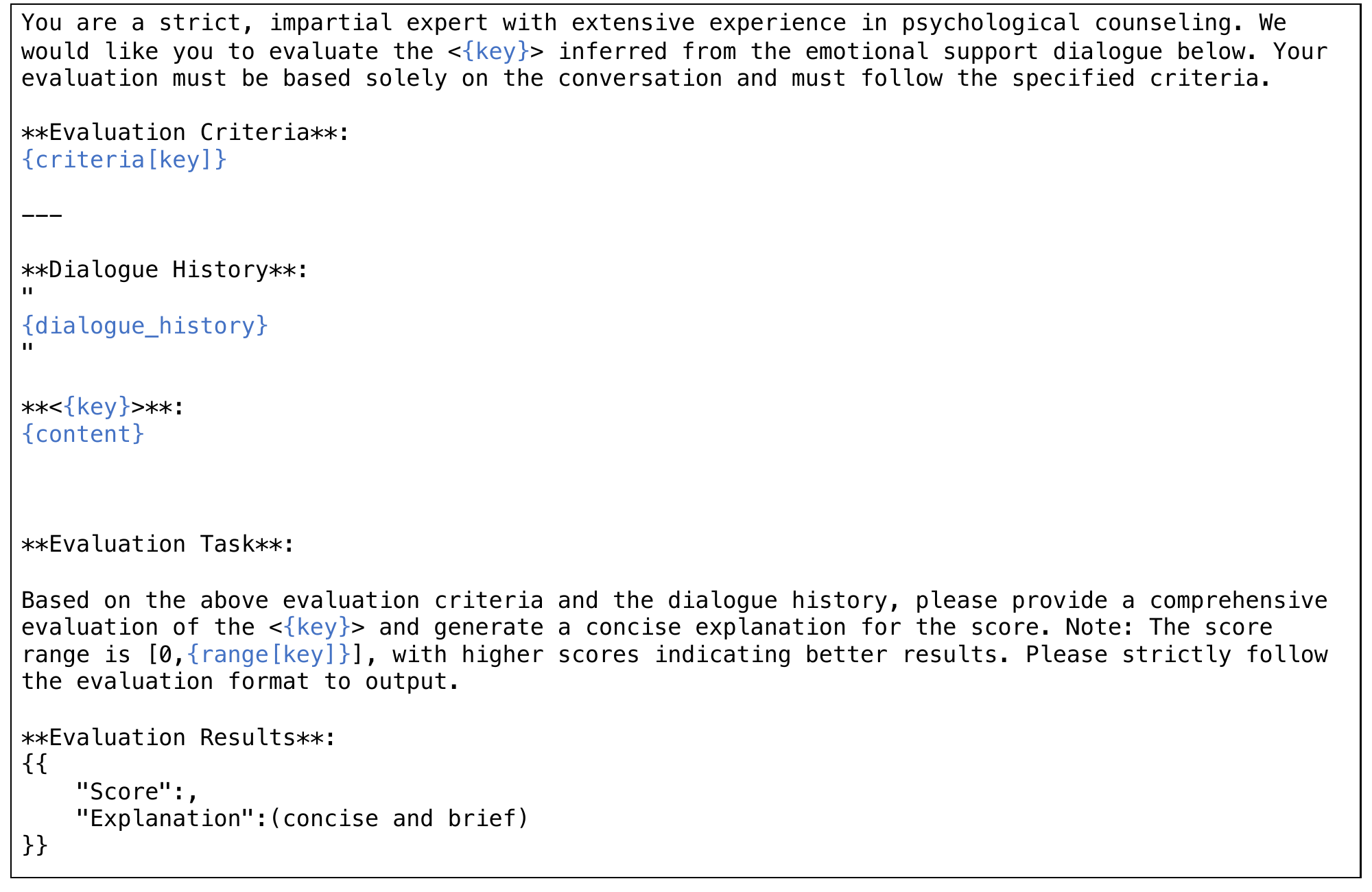}
\vspace{-8pt}
\caption{The prompt of emotional state analysis evaluation.}
\label{fig:ICECoT_ES}
\end{figure*}

(1) \textbf{Seeker's Main Issue and Underlying Causes}. The score range is [0, 4].

\begin{itemize}
    \item \textit{Accuracy} Is the <Seeker's Main Issue and Underlying Causes> exactly what the seeker mentioned in the conversation (i.e., no additional or fabricated details)?
    \item \textit{Resolution Status} Does the <Seeker's Main Issue and Underlying Causes> indicate that the seeker's issue remains unresolved, suggesting further support or discussion is needed?
    \item \textit{Inclusion of Key Information} Does the <Seeker's Main Issue and Underlying Causes> include the latest or most important details the seeker provided?
    \item \textit{Reflecting Primary Distress} Does the <Seeker's Main Issue and Underlying Causes> accurately capture the seeker's main concern or difficulty as expressed in the dialogue?
\end{itemize}

(2) \textbf{Seeker's Current Emotions and Feelings}. The score range is [0, 2].

\begin{itemize}
    \item Does the <Seeker's Current Emotions and Feelings> align with the seeker's tone, expression style, and keywords in the conversation?
    \item Does <Seeker's Current Emotions and Feelings> accurately convey the seeker's current emotional state?
\end{itemize}

(3) \textbf{Seeker's Needs}. The score range is [0, 3].

\begin{itemize}
    \item Is the <Seeker's Needs> specific and clear?
    \item Does the <Seeker's Needs> align with the seeker's expression in the conversation?
    \item Is it possible to identify implicit requirements that the seeker did not explicitly express, but can be inferred from the context or emotional flow?
\end{itemize}

(4) \textbf{Conversation Relationship Dynamics}. The score range is [0, 3].

\begin{itemize}
    \item Can the <Conversation Relationship Dynamics> accurately represent the flow of emotions and communication patterns in the interaction between the parties?
    \item Can the <Conversation Relationship Dynamics> reflect the underlying emotional tension in the conversation, showing potential conflicts, contradictions, or cooperation between the seeker and the supporter?
    \item Does the <Conversation Relationship Dynamics> display the supporter's emotional response or support methods towards the seeker, and can it demonstrate the effectiveness and quality of emotional support?
\end{itemize}

For \textbf{intention inference} evaluation, the prompt is provided in Figure~\ref{fig:ICECoT_Int}, with detailed evaluation criteria listed below:

\begin{figure*}[ht]
\centering
\includegraphics[scale=0.5]{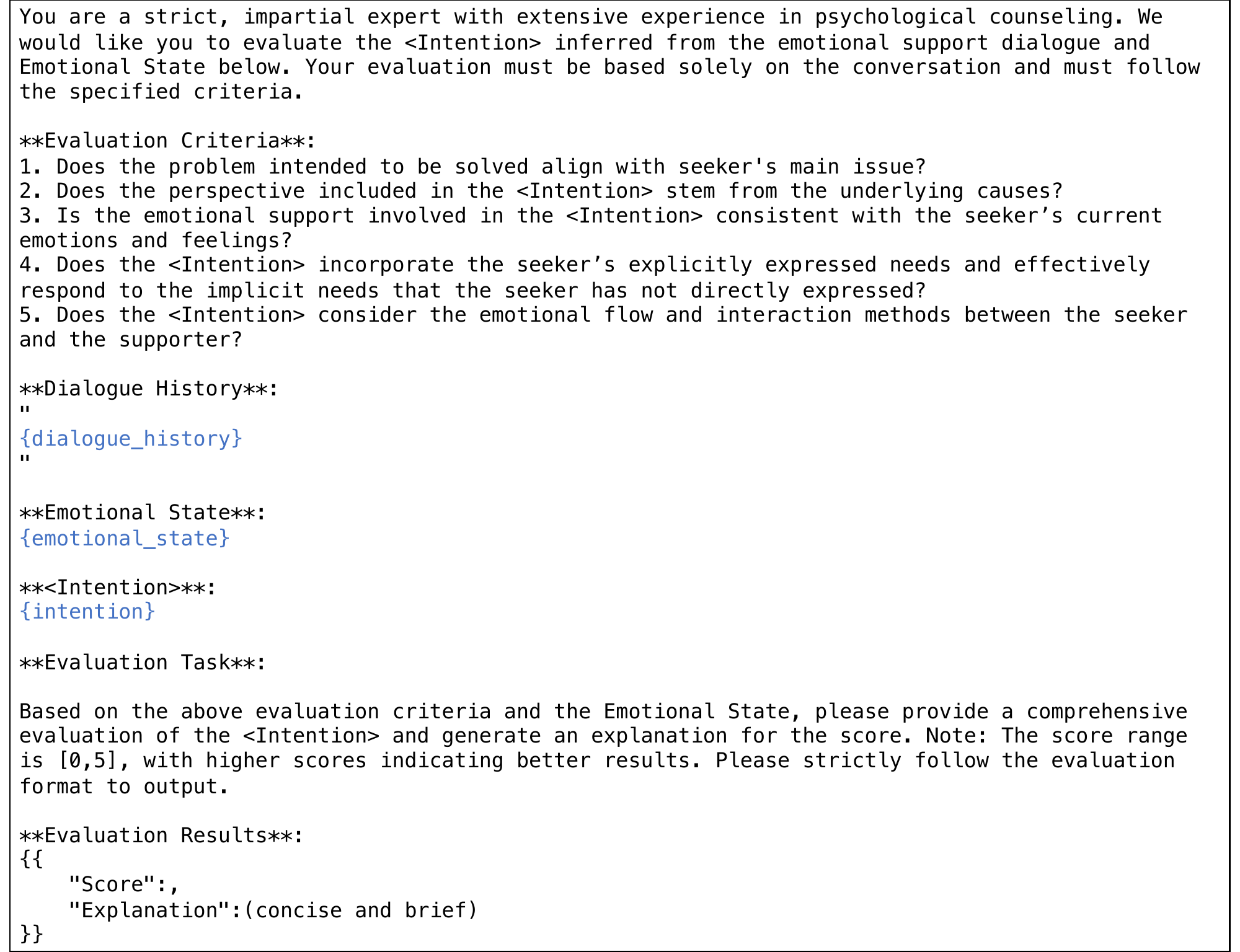}
\vspace{-8pt}
\caption{The prompt of intention inference evaluation.}
\label{fig:ICECoT_Int}
\end{figure*}

\begin{itemize}
    \item Does the problem intended to be solved align with seeker’s main issue?
    \item Does the perspective included in the <Intention> stem from the underlying causes?
    \item Is the emotional support involved in the <Intention> consistent with the seeker’s current emotions and feelings?
    \item Does the <Intention> incorporate the seeker’s explicitly expressed needs and effectively respond to the implicit needs that the seeker has not directly expressed?
    \item Does the <Intention> consider the emotional flow and interaction methods between the seeker and the supporter?
\end{itemize}

\begin{figure*}[ht]
\centering
\includegraphics[scale=0.5]{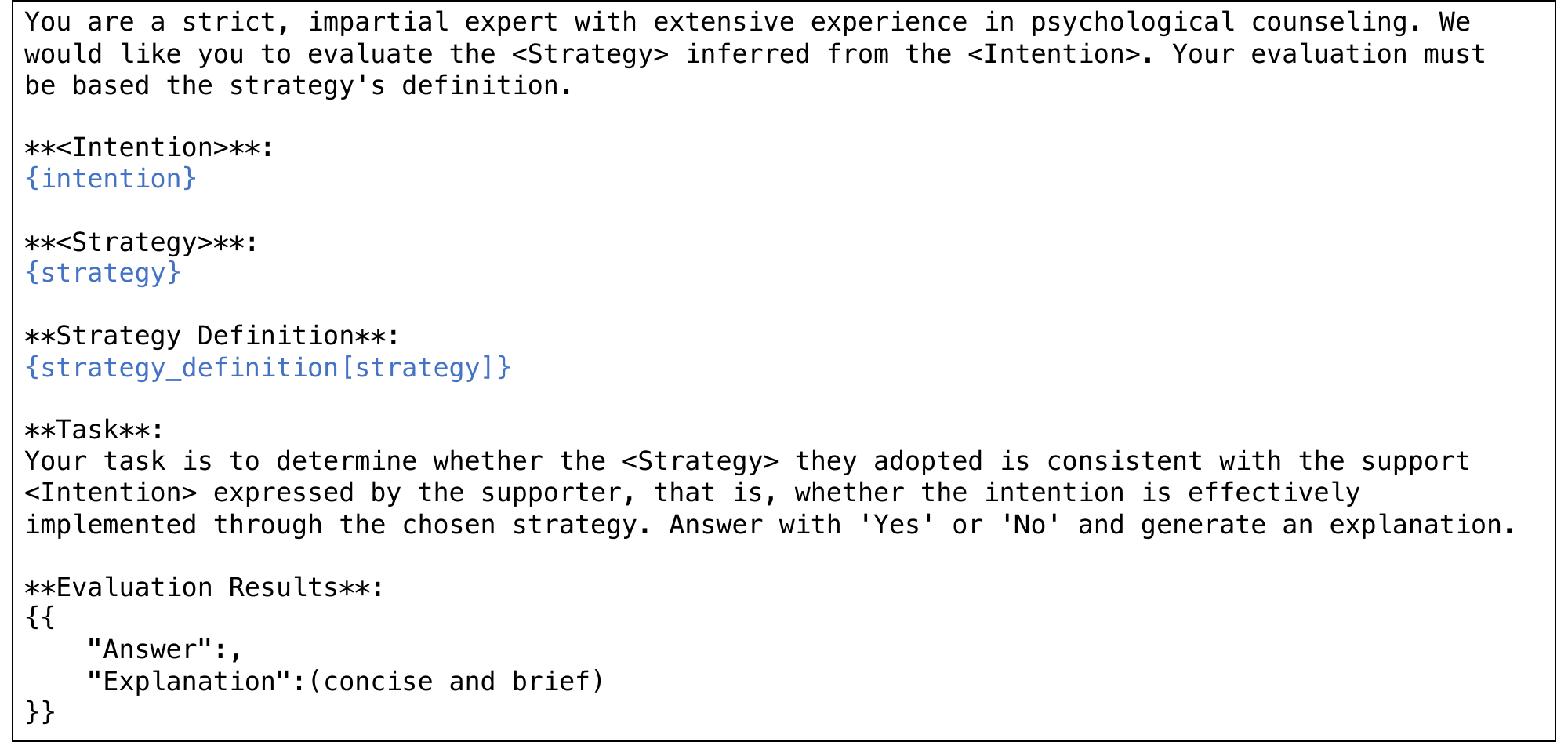}
\vspace{-8pt}
\caption{The prompt of strategy selection evaluation.}
\label{fig:ICECoT_Str}
\end{figure*}

For \textbf{strategy selection} evaluation, we incorporate strategy definitions, with the corresponding prompt displayed in Figure~\ref{fig:ICECoT_Str}.

We find that GPT-4’s performance in assessing the strategies used in responses was not satisfactory. As a result, we enlist an expert to assess the alignment between the responses and the strategies derived from the reasoning process.

\end{document}